\documentclass[lettersize,journal]{IEEEtran}
\usepackage{amsmath,amsfonts,amssymb}
\usepackage{algorithmic}
\usepackage{algorithm}
\usepackage{array}
\usepackage{caption}
\usepackage{subcaption}
\usepackage{textcomp}
\usepackage{stfloats}
\usepackage{url}
\usepackage{verbatim}
\usepackage{graphicx}
\usepackage{cite}
\usepackage{multicol}
\usepackage{color}
\def\BibTeX{{\rm B\kern-.05em{\sc i\kern-.025em b}\kern-.08em
    T\kern-.1667em\lower.7ex\hbox{E}\kern-.125emX}}

\begin{document}

\title{LAPA-based Dynamic Privacy Optimization for Wireless Federated Learning in Heterogeneous Environments}

\author{Pengcheng Sun,
Erwu Liu,
Wei Ni, \IEEEmembership{Fellow,~IEEE,}
Rui Wang,
Yuanzhe Geng,
Lijuan Lai,
and Abbas Jamalipour, \IEEEmembership{Fellow,~IEEE}
\thanks{P. Sun, E. Liu, R. Wang, Y. Geng, and L. Lai are with the College of Electronics and Information Engineering, Tongji University, Shanghai 201804, China, E-mails: pc\_sun2020@tongji.edu.cn, erwu.liu@ieee.org, ruiwang@tongji.edu.cn, yuanzhegeng@tongji.edu.cn, 2154310@tongji.edu.cn.
}
\thanks{W. Ni is with the Commonwealth Science and Industrial Research Organization (CSIRO), Marsfield, NSW 2122, Australia, E-mail: wei.ni@data61.csiro.au.
}
\thanks{This work is supported in part by grants from the National Science Foundation of China (No. 42171404, No.42225401) and Shanghai Engineering Research Center for Blockchain Applications And Services (No. 19DZ2255100).
}
\thanks{A. Jamalipour is with the School of Electrical and Information Engineering Faculty of Engineering, The University of Sydney, Sydney, NSW 2006, Australia, E-mail: a.jamalipour@ieee.org.
}
\thanks{Corresponding author: Erwu Liu.
}
}

\markboth{Journal of \LaTeX\ Class Files,~Vol.~14, No.~8, August~2021}%
{Shell \MakeLowercase{\textit{et al.}}: A Sample Article Using IEEEtran.cls for IEEE Journals}


\maketitle

\begin{abstract}
Federated Learning (FL) is a distributed machine learning paradigm based on protecting data privacy of devices, which however, can still be broken by gradient leakage attack via parameter inversion techniques. 
Differential privacy (DP) technology reduces the risk of private data leakage by adding artificial noise to the gradients, but detrimental to the FL utility at the same time, especially in the scenario where the data is Non-Independent Identically Distributed (Non-IID). 
Based on the impact of heterogeneous data on aggregation performance, this paper proposes a Lightweight Adaptive Privacy Allocation (LAPA) strategy, which assigns personalized privacy budgets to devices in each aggregation round without transmitting any additional information beyond gradients, ensuring both privacy protection and aggregation efficiency.
Furthermore, the Deep Deterministic Policy Gradient (DDPG) algorithm is employed to optimize the transmission power, in order to determine the optimal timing at which the adaptively attenuated artificial noise aligns with the communication noise, enabling an effective balance between DP and system utility.
Finally, a reliable  aggregation strategy is designed by integrating communication quality and data distribution characteristics, which improves aggregation performance while preserving privacy. Experimental results demonstrate that the personalized noise allocation and dynamic optimization strategy based on LAPA proposed in this paper enhances convergence performance while satisfying the privacy requirements of FL.
\end{abstract}

\begin{IEEEkeywords}
Federated learning, wireless communication, differential privacy, heterogeneous environments.
\end{IEEEkeywords}

\section{Introduction}

\IEEEPARstart{T}{raditional} distributed machine learning requires collecting raw data from devices to a parameter server for centralized model training, which is a paradigm that offers high accuracy. However, in many privacy-sensitive scenarios—such as speech recognition\cite{granqvist2020improving,hard2020training}, medical diagnosis\cite{brisimi2018federated,choudhury2020predicting,sadilek2021privacy}, and financial services\cite{long2020federated,yang2019ffd}—clients are often unwilling to share local private data with any entity, including the parameter server. Federated learning (FL) requires devices to train models locally and share only training parameters, providing privacy protection and thus has found broad application.

However, the training network and the resulting model information in FL are inherently vulnerable to attacks that compromise privacy, leading to serious privacy leakage risks such as Gradient Leakage Attacks (GLA)~\cite{zhu2019deep, zhao2020idlg, yin2021see, wang2019beyond, geiping2020inverting}. In such attacks, attackers construct synthetic data to approximate the original training data by minimizing the difference between virtual gradients and shared real gradients. Other common privacy attacks include membership inference attacks~\cite{song2021systematic, salem2018ml, hui2021practical}, class representative reconstruction attacks~\cite{song2017machine, hitaj2017deep}, and property inference attacks~\cite{ganju2018property}. The work by~\cite{fredrikson2015model} demonstrated a model inversion attack that recovers images from a facial recognition system. Therefore, it is necessary to protect the privacy of each device’s data in FL systems.

Differential privacy (DP) is an effective privacy protection technique and has been widely used in FL in recent years. The work in~\cite{dwork2014algorithmic} introduced the basic theory of DP. According to this work, random noise is added to parameters to enable privacy protection in the system. Under the DP mechanism, designing a proper privacy allocation strategy is an important research topic, as it affects both the level of privacy protection for each device and the convergence performance of the FL system.

The work in~\cite{truex2019hybrid} proposed a method that combines DP with secure multi-party computation to enhance the security of FL. In~\cite{geyer2017differentially}, the authors showed that under a given privacy level, the FL system can achieve good convergence performance when the number of participating devices is large enough. However, these works only consider security or privacy during the global model update phase, while the upload phase also needs to be addressed. In addition, they do not provide a theoretical explanation of the relationship between DP and FL convergence performance. The work in~\cite{wei2020federated} considers both the upload phase and the global update phase, and mathematically explains the impact of DP on FL convergence. \cite{zhou2022pflf} further examines the impact of the number of global iterations on DP-FL. However, most existing works that apply DP to FL do not consider the Non-Independent Identically Distributed (Non-IID) data in real scenarios, which significantly degrades the aggregation performance of FL when noise is added, and may even lead to divergence. X. You et al.~\cite{you2022reschedule} proposed a privacy-preserving aggregation scheme under time-varying data distribution. However, they did not consider that the privacy protection requirement may vary during the FL process, resulting in excessive redundant noise that reduces aggregation performance. The work in~\cite{sun2024socially} proposed a dynamic privacy budget allocation strategy by quantifying the Risk of Privacy Leakage (RoPL), but the algorithm requires devices to upload local test results, increasing communication overhead. Moreover, it actively launches additional gradient leakage attacks to determine the required privacy budget level, which leads to low computational efficiency. The strategy in~\cite{zhang2021privacy} improves computational efficiency, but offers limited customization for Non-IID data. In summary, there is still a lack of lightweight and customized privacy allocation strategies for FL systems under Non-IID scenarios.

In wireless FL with DP, communication quality is another important factor affecting learning performance. On the one hand, the impact of channel noise in practical communication environments introduces inherent interference to the transmitted model, which can partially function as the artificial noise in DP. In extreme cases, it may even allow the system to meet DP requirements directly. However, convergence performance under such high-noise conditions is usually poor. Under normal environmental noise, some studies attempt to balance wireless communication and privacy performance. The work in~\cite{lin2023joint} proposed a DP-protected FL resource allocation scheme for multi-cell networks, which minimizes the total privacy leakage by jointly optimizing device association, transmission power, and DP noise power, which balances privacy and communication efficiency. The work in~\cite{wang2024p2cefl} proposed a privacy-preserving and communication-efficient FL (P2CEFL) algorithm based on sparse gradients and jittered quantization to reduce communication cost under DP guarantees. Most existing works that aim to balance communication performance and privacy focus on improving communication efficiency under privacy constraints, but they do not address the aggregation quality degradation caused by DP noise. To our knowledge, the only work that aims to improve aggregation performance under privacy constraints by enhancing communication quality is~\cite{nguyen2023time}, which jointly optimizes DP noise levels and user transmission power to balance privacy protection and communication quality. However, this work does not consider the Non-IID data, and its method of balancing communication and privacy is limited, as it requires artificial noise to be added throughout the entire training process to meet DP, which overlooks the potential contribution of environmental noise. Therefore, under Non-IID conditions, an efficient privacy aggregation scheme is needed that adapts to FL iterations and takes both communication quality and data heterogeneity into account. On the other hand, under Non-IID conditions, jointly considering communication quality to improve aggregation performance is also important.

Inspired by the above works on privacy solutions, this paper proposes the Lightweight Adaptive Privacy Allocation (LAPA) strategy based on the dynamic impact of heterogeneous data on aggregation performance. The LAPA strategy is lightweight and requires no additional transmission beyond the local gradients already needed by the FL algorithm, which helps reduce network communication load and improve computational efficiency. Then, we propose a dynamic noise control mechanism by fully utilizing the potential DP disturbance function of environmental noise. When the level of adaptively attenuated artificial DP noise aligns with the impact of communication noise, we stop adding extra artificial noise to the system, as the inherent environmental noise is already sufficient to meet the DP requirement. The optimal balance point between artificial noise and communication noise is determined by optimizing the transmission power, which achieves the best trade-off between DP protection and system utility. Finally, the Wasserstein distance is used to quantify the Non-IID degree of data distribution across devices, and a user selection mechanism based on devices' SNRs is designed to determine the aggregation weights, which further ensures aggregation performance while preserving privacy.

In summary, the main contributions of this paper are as follows:

\begin{itemize}
	\item This paper designs a new LAPA algorithm. To address the Non-IID nature of data distribution and the evolving RoPL during global updates, an efficient privacy allocation strategy is proposed based on global gradient updates and local gradient contributions.
	
	\item This paper fully utilizes the distribution characteristics of data and communication capabilities across devices to assign customized aggregation weights, maximizing system aggregation performance under privacy constraints.
	
	\item This paper designs a new dynamic noise control mechanism. Specifically, before the $T_{{th}}$-th global update, the proposed LAPA strategy is used to calculate the required artificial noise to satisfy DP; after the $T_{{th}}$-th update, the inherent environmental noise is used to meet the DP requirement.
	
	\item This paper analyzes the convergence upper bound of wireless FL under the proposed noise allocation mechanism. It provides both privacy guarantees and convergence guarantees of the proposed strategy, and further formulates a transmission power optimization problem based on the obtained convergence bound, determining the optimal switching time $T_{{th}}$ via the Deep Deterministic Policy Gradient (DDPG) algorithm for noise allocation to maximize the aggregation utility of wireless FL.
\end{itemize}

The remainder of this paper is structured in the following manner: Section II describes the system model, including the learning model, the wireless communication model, and the privacy threat model. Section III describes the efficient privacy aggregation strategy based on proposed LAPA algorithm. Based on the LAPA algorithm, Section IV describes the proposed dynamic noise control optimization and provides a convergence analysis to formulate the optimization problem with respect to the transmit power.  Section V describes the simulations.  Section VI provides the conclusions. 

\textit{Notation}: Upper- and lower-case boldface letters denote matrices and vectors, respectively; $\mathbb{R}^n$ denotes the $n$-dimensional real vector space; $\mathbb{C}^{n_1 \times n_2}$ denotes the $n_1 \times n_2$-dimensional complex space;  $\left|\cdot \right|$ denotes modulus; $\|\cdot\|$ denotes Euclidean norm; $\nabla(\cdot)$ and $\nabla^2(\cdot)$ take gradient and Laplacian, respectively; $diag\left( \cdot \right)$ stands for a diagonal matrix; $\mathbb{E}\left(\cdot\right)$ takes mathematical expectation; $\left(\cdot\right)^\top$ and $\left(\cdot\right)^H$ stand for transpose and conjugate transpose, respectively.

\section{System Model}

We consider an FL system consisting of a BS equipped with $N_a$ antennas, serving as the parameter server, and $K$ single-antenna devices. The $k$-th device ($k = 1, 2, \ldots, K$) has its own local data set $\mathcal{D}_k$. Consider an FL algorithm with the input data vector $\boldsymbol{x}_{ks} \in \mathbb{R}^d$ and the corresponding output $y_{ks} \in \mathbb{R}$, where $s\in \{1,\cdots,|\mathcal{D}_k|\}$ is the index of a data sample. Let $\boldsymbol{w}_k$ denote the local model parameters trained on the $k$-th device.

\subsection{Learning Model}

For edge devices, the goal of local training is to find the optimal model $\boldsymbol{w}^\ast$ that minimizes the training loss. Without loss of generality, the model parameter $\boldsymbol{w}\in\mathbb{R}^q$ (where $q$ denotes the model size) has its local gradient with respect to dataset $\mathcal{D}_k$ at the $t$-th communication round defined as
\begin{equation}
	\nabla F_k\left(\boldsymbol{w}^{[t]}\right)=\frac{1}{|\mathcal{D}_k|}\sum_{\left(\boldsymbol{x}_{ks},y_{ks}\right)\in\mathcal{D}_k} \nabla f_k\left(\boldsymbol{x}_{ks},y_{ks};\boldsymbol{w}^{[t]}\right)\label{gradient}
	,\end{equation}
where $f_k\left(\boldsymbol{x}_{ks},y_{ks};\boldsymbol{w}\right)$ denotes the sample loss.

To minimize the global loss function, FL proceeds through multiple rounds of gradient/parameter transmission and iteration until convergence. In each communication round, the BS aggregates gradients received from $K$ users as follows:
\begin{equation}
	\nabla F\left(\boldsymbol{w}^{[t]}\right)=\sum_{k=1}^{K}G_k \cdot \nabla F_k\left(\boldsymbol{w}^{[t]}\right),
	\label{aggregationw}
\end{equation}
where $\sum_{k=1}^{K}G_k = 1$ represents the aggregation weights. Typically, when using the FedAvg algorithm, the aggregation weight is set as $G_k = |\mathcal{D}_k| / \sum_{k=1}^K{|\mathcal{D}_k|}$. However, under Non-IID conditions, this aggregation strategy clearly fails to achieve efficient convergence. Therefore, $G_k$ needs to be redesigned based on practical scenarios, which will be detailed in Section III-C.

Finally, the global model is updated as
\begin{equation}
	\boldsymbol{w}^{[t+1]}= \boldsymbol{w}^{[t]} - \lambda \sum_{k=1}^{K} G_k \cdot \nabla F_k\left(\boldsymbol{w}^{[t]}\right),
	\label{global}
\end{equation}
where $\lambda$ denotes the learning rate.

\subsection{Communication Model}

The transmission of local gradients and global parameters between devices and the BS is carried out through a wireless communication system. The NOMA framework used in this paper allows multiple devices to transmit data simultaneously over a superimposed wireless channel, effectively alleviating communication congestion caused by multiple devices in FL.

A block fading channel model is assumed, where the channel coefficients remain constant throughout the training process. Let $\boldsymbol{h}_k \in \mathbb{C}^{N_a \times 1}$ denote the channel coefficient vector from the $k$-th device to the BS, with magnitude modeled as an independent random variable. It is assumed that perfect channel state information is available at both the BS and the devices. Therefore, at the $t$-th aggregation round, the signal received at the BS is given by:
\begin{equation}
	\boldsymbol{y}^{[t]} = \sum_{k=1}^{K} p_k \boldsymbol{h}_k \boldsymbol{s}_k^{[t]} + \boldsymbol{n}_0,
\end{equation}
where $p_k \in \mathbb{C}$ is the transmit power scalar of the $k$-th device, $\boldsymbol{s}_k$ is the gradient vector transmitted from the device to the BS, and $\boldsymbol{n}_0 \in \mathbb{C}^{N_a \times q}$ is the Additive White Gaussian Noise (AWGN) with each element following a distribution $\mathcal{CN}(0, \sigma_{n_0}^2)$, where $\sigma_{\boldsymbol{n}_0}^2$ denotes the environmental noise power.

The transmit power of the $k$-th device is constrained by:
\begin{equation}
	\mathbb{E}\left( |p_k s_k|^2 \right) = |p_k|^2 \leq P_0,
\end{equation}
where $P_0 > 0$ is the upper bound of the maximum transmit power.

The Serial Interference Cancellation (SIC) technique based on least squares estimation is adopted to address transmission interference. Therefore, the received Signal-to-Interference-plus-Noise Ratio (SINR) of the $k$-th device is:
\begin{equation}
	\gamma_k=\frac{p_{k}^2\left|\boldsymbol{r}_{k}\boldsymbol{h}_{k}\right|^2}{\sum_{i=k+1}^K p_{i}^2\left|\boldsymbol{r}_{k}\boldsymbol{h}_{i}\right|^2+\|\boldsymbol{r}_{k}\|^2 \sigma_n^2},
\end{equation}
where $\boldsymbol{r}_k$ is the linear receiver. To explicitly reflect the impact of environmental noise on the transmission system, we assume that the SIC technique can fully cancel inter-device interference and let $\boldsymbol{r}_k = \boldsymbol{h}_k^{H}$. As a result, the BS decodes the gradient signal from the $k$-th device as $\hat{\boldsymbol{s}}_k = \boldsymbol{s}_k + \frac{\sigma_{n_0}}{p_k \|\boldsymbol{h}_k\|}$, where $\frac{\sigma_{\boldsymbol{n}_0}}{p_k \|\boldsymbol{h}_k\|}$ denotes the influence of environmental noise on the transmission system.

\subsection{Threat Model}

In the FL system, we assume that all participants other than the devices are honest-but-curious, especially the BS acting as the parameter server, which means that the attacker follows the training protocol honestly but is curious about the private data of target devices. The attacker always attempts to recover private data during the training or testing phase by reversing the shared model parameters between devices and the BS. Therefore, even though FL allows training and storing private data locally, there is still a RoPL, which is a common assumption in most studies analyzing potential privacy risks in FL. Specifically, this paper assumes that the attacker may possess a dataset $\mathcal{D}_{{attack}}$ that overlaps with the data of some participating devices. The attacker then tries to use a subset $\mathcal{D}_{{attack}}'$ to simulate the local training of the $k$-th device, and infers private data of the device by exploiting the difference in model outputs between training and non-training samples. Hence, this model is realistic and reasonable.

\section{Efficient Privacy Aggregation based on LAPA}

\subsection{Definition of DP}

To ensure the privacy of algorithms on datasets, the DP mechanism provides a strong guarantee through the parameters $\epsilon$ and $\delta_{dp}$. Suppose the dataset $\mathcal{D}$ contains neighboring datasets $d$ and $d'$, which differ by only one sample. $\epsilon > 0$ defines the distinguishability bound for all outputs of the mapping $\mathcal{M}$ when applied to neighboring datasets $d$ and $d'$. A larger value of $\epsilon$ indicates a higher distinguishability between the neighboring datasets, and thus a greater RoPL. The parameter $\delta_{dp}$ is not part of the original DP definition, but in the extended version proposed by Dwork et al.~\cite{dwork2006our}, it allows relaxation of the strict $\epsilon$-DP requirement, making DP more flexible in practice. $\delta_{dp}$ represents the probability that the output difference between two neighboring datasets is not bounded by $e^{\epsilon}$ after applying the privacy mechanism, i.e., the probability of DP failure. The definition of $(\epsilon,\delta_{dp})$-DP is as follows.

\vspace{2 mm}
\noindent\textbf{Definition 1 ($( \epsilon,\delta_{dp} )$}-DP~\cite{wang2019local}: \textit{A randomized mechanism $\mathcal{M}: \mathcal{D} \rightarrow \mathcal{R}$, with domain $\mathcal{D}$ and range $\mathcal{R}$, satisfies $(\epsilon,\delta_{dp})$-DP if for any two adjacent inputs $d, d' \in \mathcal{D}$ and any output $\mathcal{S} \in \mathcal{R}$, the following holds:
	\begin{equation}
		\Pr[\mathcal{M}(d) \in \mathcal{S}] \leq e^{\epsilon} \Pr[\mathcal{M}(d') \in \mathcal{S}] + \delta_{dp}.
	\end{equation}
	This ensures that the output probabilities for neighboring datasets are similar within a bounded range, and the deviation is controlled by $\delta_{dp}$.}
\vspace{2 mm}

A common method to approximate a deterministic real-valued function $\mathcal{M}: \mathcal{D} \rightarrow \mathcal{R}$ under the DP mechanism is to add Gaussian noise with zero mean and variance $\sigma^2 \mathbf{I}$ to each coordinate of the function output $s(d)$, where $\mathbf{I}$ is the identity matrix of the same dimension as $s(d)$. This process is expressed as:
\begin{equation}
	\mathcal{M}(d) = s(d) + \mathcal{N}(0, \sigma^2 \mathbf{I}).
\end{equation}

The $l_2$-norm-based sensitivity is defined as:
\begin{equation}
	\Delta s = \max_{d, d' \in \mathcal{D}} \left\| s(d) - s(d') \right\|_2,
\end{equation}
which provides an upper bound on the amount of perturbation that needs to be added to the output for privacy protection. Repeated application of this noise-adding DP mechanism can be implemented using the basic composition theorem, the advanced composition theorem, or their improvements, which enables devices to design personalized DP guarantees.

\subsection{LAPA Algorithm}

\begin{figure*} 
\centerline{\includegraphics[width=0.8\linewidth]{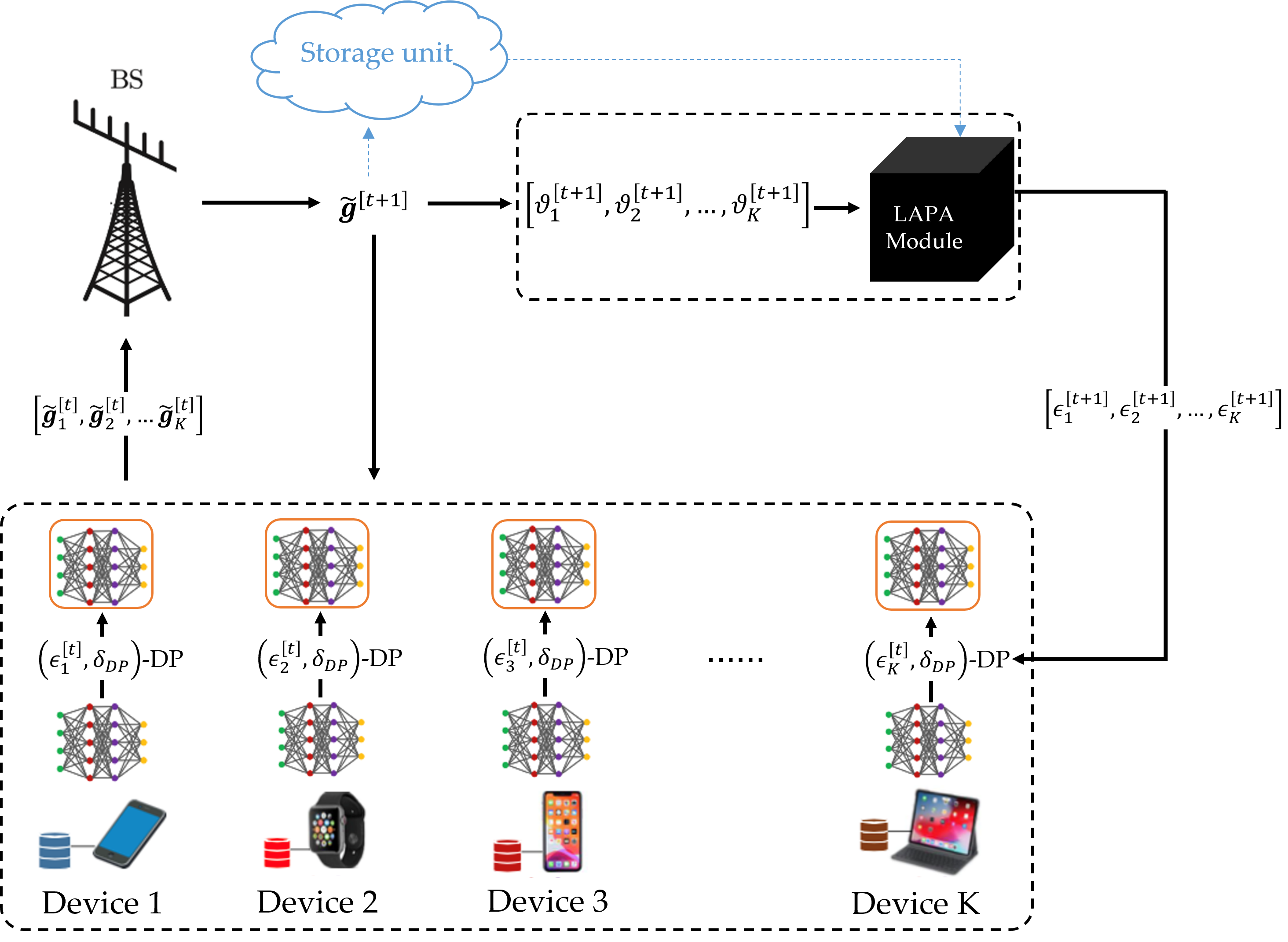}}
\caption{The workflow of proposed LAPA-based FL, involving only gradient exchange between devices and the BS, without incurring additional communication overhead, making it simpler and more efficient than the algorithm in~\cite{hu2023shield}.
}
\label{LAPAworkflow}
\end{figure*}

To satisfy the DP requirement of wireless FL, this paper adopts the $(\epsilon,\delta_{dp})$-DP criterion to apply noise perturbation to the gradients to be transmitted. In learning systems with Non-IID data condition, applying a uniform level of artificial noise to all devices can meet the privacy requirement, but it severely degrades the convergence performance of the system due to the significant data distribution differences and additional noise disturbances. Therefore, it is necessary to design a personalized privacy allocation scheme. Fig.~\ref{LAPAworkflow} illustrates the wireless FL framework based on LAPA proposed in this paper. In the $t$-th global aggregation round, heterogeneous devices train local model gradients based on their local data, compute customized artificial noise power according to the designed privacy budget allocation strategy, and add noise to each element of the gradient before transmitting it to the BS for aggregation. This process is repeated until the system converges. The following section introduces the privacy allocation algorithm in detail.

Section III-C of~\cite{hu2023shield} points out that the RoPL changes with the progress of FL. Specifically, in the early stages of training, devices face a higher RoPL, while this risk decreases rapidly as the training process converges, following an approximately exponential trend. Therefore, the privacy budget in each FL iteration can be allocated according to the training progress. As a result, quantifying the progress of FL training is important.

The training progress of FL can be quantified by the change in the global gradient over communication rounds. Specifically, in the $t$-th communication round, the BS aggregates the locally perturbed gradients $\tilde{\boldsymbol{g}}_k^{[t]} = \nabla F_k(\tilde{\boldsymbol{w}}^{[t]})$ to form the global perturbed gradient $\tilde{\boldsymbol{g}}^{[t]} = \nabla F(\tilde{\boldsymbol{w}}^{[t]})$, which is stored in a memory unit. Then, the “difference” in the global gradient relative to its historical state is used as an indicator to measure training progress (i.e., whether convergence is approaching), which helps quantify the RoPL at each communication round.

A variant of the Proportional-Integral-Derivative (PID) algorithm~\cite{wang2016rescuedp} is used to describe the dynamic changes of the global gradient, as it is highly sensitive to data and effectively guides the allocation of the privacy budget $\epsilon^{[t]}$ for each communication round, even when the changes are small. Compared to the work in~\cite{hu2023shield}, the advantage of using the global gradient to reflect training progress in this paper is that no additional information needs to be transmitted—such as accuracy on a public validation set—other than the local gradients already required by FL, which significantly reduces communication overhead. The storage unit on the BS records the global gradient information of round $t$ and its historical updates as $\nabla F(\tilde{\boldsymbol{w}}) = \left[ \nabla F(\tilde{\boldsymbol{w}}^{[1]}), \nabla F(\tilde{\boldsymbol{w}}^{[2]}), \dots, \nabla F(\tilde{\boldsymbol{w}}^{[t-1]}) \right]$. From the interval $[t - m, t]$ in $\nabla F(\tilde{\boldsymbol{w}})$, the BS randomly selects $m$ samples, where the index of the current sample is $i_m$ and the adjacent sample is $i_{m-1}$, to compute the PID-like error. The feedback error between the current sample and its adjacent sample in terms of the global gradient is given by:
\begin{equation}
	E^{(i_m)} = \left| \nabla F\left( \tilde{\boldsymbol{w}}^{[i_m]} \right) - \nabla F\left( \tilde{\boldsymbol{w}}^{[i_{m-1}]} \right) \right|.
\end{equation}
Then, the PID-like error of round $t$ on the sampled subset of $\nabla F(\tilde{\boldsymbol{w}})$ is calculated as:
\begin{equation}
	e^{[t]} = K_p E^{(i_m)} + K_s \left( \frac{1}{m} \sum_{s = n - m + 1}^{m} E^{(i_s)} \right),
\end{equation}
where $K_p$ and $K_s$ denote the proportional and integral coefficients, respectively.

Let the total privacy budget of the system be $\epsilon^T$, which is first allocated to each communication round, resulting in $\epsilon^{[t]}$. We require that $\epsilon^{[t]}$ be monotonically non-decreasing with respect to the RoLP, i.e., monotonically non-increasing with respect to the training progress, and that their total sum does not exceed the total privacy budget $\epsilon^T$. Let $\epsilon_c = \sum_{i=1}^{t-1} \epsilon^{[t]}$ denote the privacy budget consumed up to the $t$-th round. The remaining privacy budget is allocated to each communication round based on the nearly exponential trend of RoLP with respect to FL progress, as follows:
\begin{equation}
	\epsilon^{[t]} = \exp(-e^{[t]}) \cdot \frac{\epsilon^T - \epsilon_c}{T - t + 1}, \quad s.t. \sum_{t=1}^{T} \epsilon^{[t]} \leq \epsilon^T.
\end{equation}
Next, to meet the personalized privacy requirements of devices with heterogeneous characteristics and to better balance FL aggregation performance and privacy utility, we further allocate $\epsilon^{[t]}$ to each heterogeneous device. Data distribution heterogeneity leads to a deviation between local and global gradients, meaning that devices exhibit heterogeneity in their local gradient updates. Therefore, to achieve better learning performance under privacy constraints, we design a personalized privacy budget allocation scheme based on the data heterogeneity of devices. In the $t$-th round, the deviation of device $k$’s local gradient from the global gradient is expressed as:
\begin{equation}
	\vartheta_k^{[t]} = \arccos \left( \frac{ \left\langle \nabla F(\tilde{\boldsymbol{w}}^{[t]}),\ \nabla F_k(\tilde{\boldsymbol{w}}^{[t]}) \right\rangle }
	{ \left\| \nabla F(\tilde{\boldsymbol{w}}^{[t]}) \right\| \cdot \left\| \nabla F_k(\tilde{\boldsymbol{w}}^{[t]}) \right\| } \right),
\end{equation}
where a smaller $\vartheta_k^{[t]}$ indicates a smaller “difference” between the device’s local gradient and the global gradient in the current round, meaning the device contributes more to the global update and should receive stronger privacy protection. To suppress the instability of random angles, we smooth the angle between round $t$ and round $t-1$ as follows:
\begin{equation}
\tilde{\vartheta}_k^{[t]} =
\begin{cases}
	\vartheta_k^{[t]}, & t = 1 \\
	\frac{t-1}{t} \tilde{\vartheta}_k^{[t-1]} + \frac{1}{t} \vartheta_k^{[t]}, & t > 1
\end{cases}.
\end{equation}
Then, the following mapping function is used to quantify the dynamic contribution of the $k$-th device in round $t$:
\begin{equation}
	f(\tilde{\vartheta}_k^{[t]}) = \beta \left( 1 - \exp \left( - \exp \left( -\beta(\tilde{\vartheta}_k^{[t]} - 1) \right) \right) \right),
\end{equation}
where $\beta$ is the decay factor. This mapping function is a monotonically non-decreasing function with respect to the smoothed angle. A smaller smoothed angle indicates that the device’s local gradient is closer to the global gradient in the current round, representing a higher contribution. Therefore, this function can be used to measure the dynamic contribution of devices with respect to data distribution.

The learning system in this paper is under the Non-IID condition. Therefore, the dynamic contributions of devices are generally independent in the $t$-th round. Accordingly, the privacy budget $\epsilon^{[t]}$ of round $t$ can be personalized and allocated to each device based on the relative magnitude of their dynamic contributions, serving as guidance for the next update round:
\begin{equation}
	\epsilon_k^{[t+1]} = \frac{f(\tilde{\vartheta}_k^{[t]})}{\sum_{k=1}^{K} f(\tilde{\vartheta}_k^{[t]})} \cdot \epsilon^{[t]}, 
	\quad {s.t.} \quad \sum_{k=1}^{K} \epsilon_k^{[t+1]} \leq \epsilon^{[t]}.
\end{equation}

The advantages of the LAPA algorithm proposed in this paper are as follows: 1) The privacy budget allocation in~\cite{zhang2021privacy} is relatively coarse, simply increasing or decreasing the next round's privacy budget without capturing the heterogeneity among devices. In fact, it is not a truly personalized privacy budget allocation scheme. In contrast, our method achieves personalization based on the measurement of data heterogeneity; 2) The user-level privacy budget allocation algorithm in~\cite{hu2023shield} is overly complex, requiring manual simulation of gradient leakage attacks in each round, which reduces computational efficiency. Our algorithm is much simpler and does not require manually simulating gradient leakage attacks.

In this paper, the parameter server is assumed to be semi-honest. Therefore, to prevent eavesdropping by third parties and potential leakage by the server, we directly address data and privacy heterogeneity among devices~\cite{li2020federated}. In other words, we focus on privacy performance in the uplink phase. At the same time, this customized privacy protection mechanism helps achieve better model accuracy and privacy guarantees, as confirmed by simulation results.

To satisfy DP guarantees, each local gradient must be clipped before adding artificial noise. The clipping is performed under the $l_2$-norm as follows:
\begin{equation}
	\boldsymbol{g}_k^{[t]} = \frac{\boldsymbol{g}_k^{[t]}}{\max\left(1,\ \left\| \boldsymbol{g}_k^{[t]} \right\| / C\right)},
\end{equation}
where $C$ is the clipping threshold. This ensures that when $\left\| \boldsymbol{g}_k^{[t]} \right\| \leq C$, $\boldsymbol{g}_k^{[t]}$ is retained as is; otherwise, its norm is constrained within $C$. This helps reduce the impact of added noise on the gradient, mitigating its negative effect on training updates while still satisfying DP. Accordingly, the $l_2$-norm sensitivity of the local gradient can be expressed as:
\begin{equation}
    \begin{aligned}
    	\Delta s &= \max_{\mathcal{D}_k,\ \mathcal{D}_k'} \left\| s^{\mathcal{D}_k} - s^{\mathcal{D}_k'} \right\| \\
	&= \max_{\mathcal{D}_k,\ \mathcal{D}_k'} \left\| \arg \min_{\boldsymbol{w}_k} \left[ \boldsymbol{g}_k^{\mathcal{D}_k} \right] 
	- \arg \min_{\boldsymbol{w}_k} \left[ \boldsymbol{g}_k^{\mathcal{D}_k'} \right] \right\| \\
    &= \frac{2\lambda C}{|\mathcal{D}_k|}.
    \end{aligned}
\end{equation}

Finally, before uploading to the BS, noise is added as follows:
\begin{equation}
	\tilde{\boldsymbol{g}}_k^{[t]} = \boldsymbol{g}_k^{[t]} + \boldsymbol{\eta}_k^{[t]},
\end{equation}
where $\boldsymbol{\eta}_k$ is an AWGN vector following a complex Gaussian distribution $\mathcal{CN}(0,\sigma_k^2)$. The noise scale $\sigma_k^2$ satisfies $\sigma_k^{[t]} \geq \frac{c \Delta s_k}{\epsilon_k^{[t]}}$, where the constant $c \geq \sqrt{2 \ln \left( \frac{1.25}{\delta_{dp}} \right)}, \quad \epsilon_k^{[t]} \in (0, 1)$. Typically, $\delta_{dp}$ represents the probability of events where the output difference between two adjacent datasets is not bounded by $e^{\epsilon_k^{[t]}}$ after applying the privacy protection mechanism. Given any $\delta_{dp}$, a larger $\epsilon_k^{[t]}$ leads to clearer distinguishability between neighboring datasets, resulting in a higher RoPL.

\subsection{Aggregation Weights with Heterogeneity}

In this section, under the premise of ensuring system-level privacy protection, we propose an aggregation strategy that addresses both data and communication heterogeneity. The aggregation weights $G_k$ are determined using the Wasserstein distance and a device selection mechanism based on SINR. Compared with the traditional FedAvg method, which assigns weights solely based on data volume, our aggregation strategy achieves more efficient and accurate convergence under system heterogeneity.

The Wasserstein distance can be used to measure the similarity between two probability distributions. Compared with Kullback-Leibler (KL) divergence and Jensen-Shannon (JS) divergence, it offers better smoothness, as it can address the gradient vanishing problem. This is because, when two distributions do not overlap or have only slight overlap, the JS divergence remains constant (log2), while the KL divergence tends to infinity. In contrast, the Wasserstein distance can still reflect the distance between the two distributions under such conditions. Therefore, using the Wasserstein distance is beneficial for generating stable device aggregation weights in heterogeneous wireless FL systems.

The Wasserstein distance between two distributions can be computed using the probability mass functions (PMFs) ${pmf}_k$ and ${pmf}_G$, and is defined as follows:
\begin{equation}
	W_k({pmf}_k,\ {pmf}_G) = \inf_{\Gamma \in \Pi({pmf}_k,\ {pmf}_G)} \mathbb{E}_{(x,y) \sim \Gamma} \left[ \| x - y \| \right],
\end{equation}
where ${pmf}_k$ represents the PMF of the label distribution in the local dataset of device $k$, ${pmf}_G$ represents the PMF of the global dataset labels, and $\Gamma$ denotes the set of all possible joint distributions. This formulation allows us to quantify the data heterogeneity of each device.

Based on the quantification of device data heterogeneity, we further compute the aggregation weight of device $k$ when selected for training using the Softmax function:
\begin{equation}
	G_k = \frac{|\mathcal{D}_k| \cdot e^{1 / W_k}}{\sum_{k=1}^{K} |\mathcal{D}_k| \cdot e^{1 / W_k}}.
\end{equation}
This provides a stable and smooth aggregation weight. At the same time, the weight assignment allows relatively homogeneous devices to contribute more to the global update, while relatively heterogeneous devices contribute less. This helps balance the influence of each device on the global model under data heterogeneity, thereby improving the convergence and overall performance of FL.

In addition to data heterogeneity, devices also exhibit significant differences in communication capabilities, which further affects the convergence efficiency of FL, as transmission errors can severely degrade the quality of global model updates. Therefore, device selection based on communication heterogeneity is critical. Accordingly, we introduce a communication-aware device selection mechanism based on the SINR, computed by $\gamma_k$, to determine whether a device participates in training. The aggregation weight of a device is then updated as:
\begin{equation}
G_k =
\begin{cases}
	\frac{|\mathcal{D}_k| \cdot e^{1 / W_k}}{\sum_{k=1}^{K} |\mathcal{D}_k| \cdot e^{1 / W_k}}, &  \gamma_k \geq \gamma_{{th}} \\
	0, &  \gamma_k < \gamma_{{th}}.
\label{6 G_k}
\end{cases}
\end{equation}
If the threshold is set too high, the BS will receive only a small number of local models with little or no error, resulting in insufficient training. If the threshold is set too low, local models with large errors may be accepted and used in global aggregation, leading to accumulated training errors. Both cases degrades the accuracy of FL.

\section{Dynamic Noise Control Optimization}

\subsection{Dynamic Noise Control Mechanism}

The aforementioned LAPA algorithm introduces artificial noise throughout the entire training process of FL to ensure compliance with DP requirements. However, continuously relying on artificial noise may introduce excessive and unnecessary noise into the system, reducing convergence performance. It is worth noting that the RoPL decreases as training progresses (as discussed in Section III-B). Meanwhile, the inherent channel noise in the communication environment, represented by $\frac{\sigma_{\boldsymbol{n}_0}}{p_k \|\boldsymbol{h}_k\|}$, has the potential to satisfy DP constraints. Therefore, this section proposes a dynamic noise control mechanism that applies different noise injection strategies at different stages of training, aiming to optimize overall FL performance while ensuring privacy protection.

\begin{figure}
	\centerline{\includegraphics[width=1.0\linewidth]{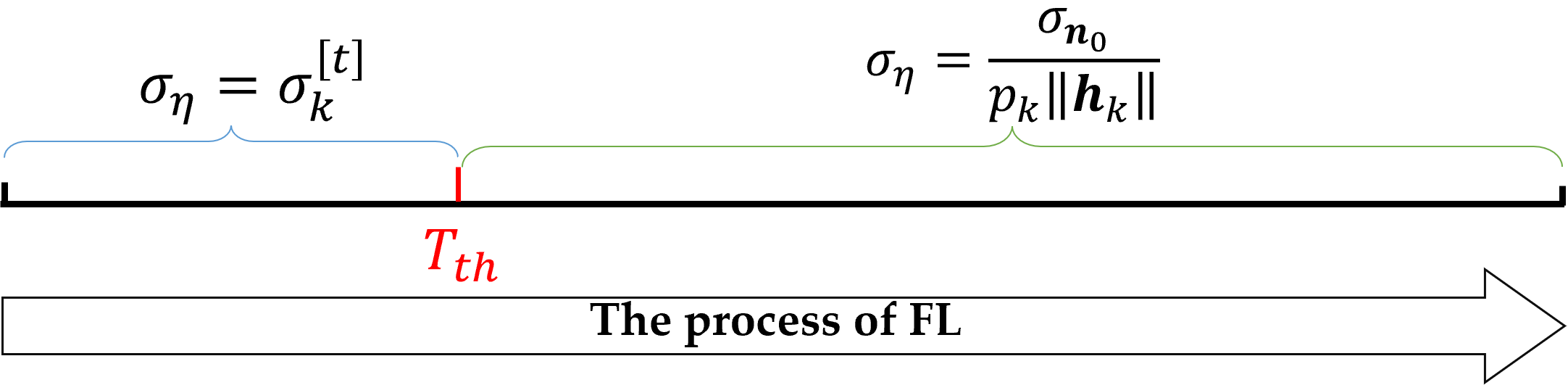}}
	\caption{dynamic noise control mechanism.}
	\label{noise}
\end{figure}

Specifically, in the early stage of training (i.e., before the $T_{{th}}$-th round of global aggregation), artificial noise is added according to the LAPA algorithm to ensure strict DP compliance and effectively suppress privacy leakage risks during the initial phase of system training. In the later stage of training (i.e., after the $T_{{th}}$-th round of global aggregation), the inherent environmental noise in the wireless channel is leveraged as a privacy-preserving mechanism to replace artificial noise. This reduces unnecessary noise interference while still ensuring that the system meets DP requirements. The proposed dynamic noise control mechanism is illustrated in Fig.~\ref{noise}.

The determination of $T_{{th}}$ depends on the timing where the required level of artificial noise decreases with training progress and aligns with the interference level of the communication environment noise, i.e.,
\begin{equation}
	\sum_{k=1}^{K} \frac{\sigma_{\boldsymbol{n}_0}}{p_k \|\boldsymbol{h}_k\|} = 
	\sum_{k=1}^{K} \frac{\Delta s_k \cdot \sqrt{2 \ln (1.25 / \delta_{dp})}}{\epsilon_k^{[T_{{th}}]}}.
\end{equation}
This equation ensures that after the $T_{{th}}$-th round of global aggregation, relying solely on environmental noise is sufficient to meet DP constraints. This noise control mechanism not only adapts dynamically to changes in privacy protection needs, but also minimizes the convergence performance degradation caused by excessive noise, while maintaining DP compliance. As a result, it improves communication efficiency and the convergence stability of the global model in FL systems.

\subsection{Convergence Analysis and Problem Formulation}

This section analyzes the convergence of FL based on the proposed LAPA algorithm and dynamic noise control mechanism. The proposed dynamic noise control mechanism leverages the combination of artificial noise and communication noise to satisfy DP constraints, where the artificial noise is generated by the LAPA algorithm and the communication interference arises from environmental noise. The transmission power of devices determines the interference intensity of communication noise, affecting the dynamic injection level of artificial noise and ultimately determining the timing where it can be reduced to match the communication interference. This process is critical to the convergence performance of FL. Therefore, it is necessary to analyze convergence to quantify the impact of transmission power on FL aggregation performance under the LAPA algorithm and the dynamic noise control mechanism.

To analyze the convergence of FL, we introduce assumptions \textbf{A1}–\textbf{A4}:

\textbf{A1} \quad $F_k$ is $L$-smooth, i.e., $\|F_k(\boldsymbol{w}) - F_k(\boldsymbol{w}')\| \leq L \|\boldsymbol{w} - \boldsymbol{w}'\|$. According to this assumption and the triangle inequality, $F(\boldsymbol{w})$ is also $L$-smooth. Additionally, its gradient is Lipschitz continuous;

\textbf{A2} \quad $F_k$ is $\mu$-strongly convex, i.e., $F_k(\boldsymbol{w}^{[t+1]}) \geq F_k(\boldsymbol{w}^{[t]}) + (\boldsymbol{w}^{[t+1]} - \boldsymbol{w}^{[t]})^\top \nabla F(\boldsymbol{w}^{[t]}) + \frac{\mu}{2} \|\boldsymbol{w}^{[t+1]} - \boldsymbol{w}^{[t]}\|^2$;

\textbf{A3} \quad $F(\boldsymbol{w})$ is second-order continuously differentiable. Based on this, together with \textbf{A1} and \textbf{A3}, we have $\mu \boldsymbol{I} \preceq \nabla^2 F(\boldsymbol{w}) \preceq L \boldsymbol{I}$;

\textbf{A4} ($\delta$-local dissimilarity) The local loss function $F_k(\boldsymbol{w}^{[t]})$ at $\boldsymbol{w}^{[t]}$ is $\delta$-locally dissimilar, i.e.,
$\mathbb{E} \left[ \|\nabla F_k(\boldsymbol{w}^{[t]})\|^2 \right] \leq \delta^2 \|\nabla F(\boldsymbol{w}^{[t]})\|^2,  \text{for } k = 1, \ldots, K$,
where $\mathbb{E}$ denotes the weighted aggregation over participating devices, and the larger the $\delta \geq 1$, the more heterogeneous the data, indicating that local updates (i.e., local gradients) are more divergent. When the data distribution is IID, $\delta$ approaches 1. Thus, $\delta$ characterizes the degree of Non-IID-ness.

In the considered FL system, image classification is taken as an example task. Accordingly, the cross-entropy function is chosen as the loss function, which is strongly convex and satisfies the aforementioned assumptions \textbf{A1}–\textbf{A4}. The following theorem provides the convergence upper bound of FL under the specified LAPA algorithm and dynamic noise control mechanism.

\vspace{2 mm}
\noindent\textbf{Theorem 1:} \textit{Given the optimal global model $\boldsymbol{w}^*$ under ideal channel conditions, the local dissimilarity index $\delta$, the aggregation weights $G_k$ computed based on the proposed aggregation strategy, the learning rate $\lambda$, and the transmission power allocation $\boldsymbol{p}$, the convergence upper bound of FL is given by:
	\begin{equation}
		\begin{aligned}
			&\mathbb{E}\left[ F\left( \boldsymbol{w}^{[t+1]} \right) - F\left( \boldsymbol{w}^* \right) \right] \\
			&\leq A^T \mathbb{E}\left[ F\left( \boldsymbol{w}^{[0]} \right) - F\left( \boldsymbol{w}^* \right) \right] \\
			&\quad + \frac{L\lambda^2}{2} \cdot \frac{1 - A^T}{1 - A} \sum_{k=1}^{K} \frac{G_k^2 \sigma_{\boldsymbol{n}_0}^2}{\|\boldsymbol{h}_k\|^2 p_k^2} \\
			&\quad + \frac{L\lambda^2}{2} \sum_{m=0}^{T_{{th}}} A^m \left[ \sum_{k=1}^{K} \frac{G_k^2 \Delta s_k^2 \cdot 2 \ln(1.25/\delta_{dp})}{(\epsilon_k^{[m]})^2} \right].
		\end{aligned}
	\end{equation}
	where $A = 1 + \mu L \lambda^2 \delta^2 \sum_{k=1}^{K} G_k^2 - 2\lambda \mu$, and $T$ is the total number of aggregation rounds.	
} 

\textit{Proof:} See Appendix I.
\vspace{2 mm}

According to \textbf{Theorem 1}, when $A < 1$, the upper bound $\mathbb{E}\left[ F(\boldsymbol{w}^{[t+1]}) - F(\boldsymbol{w}^*) \right]$ converges at a rate of $A$. Therefore, $A$ can be regarded as an indicator of the FL convergence rate. To ensure convergence of FL, i.e., to satisfy $A < 1$, the following condition should be held:
\begin{equation}
	A = 1 + \mu L \lambda^2 \delta^2 \sum_{k=1}^{K} G_k^2 - 2 \lambda \mu < 1.
\end{equation}
Based on assumption \textbf{A3}, we have $\mu < L$ (or $\mu / L < 1$). To satisfy $A < 1$, the following inequality must be satisfied:
\begin{equation}
	\mu L \lambda^2 \delta^2 \sum_{k=1}^{K} G_k^2 < 2 \lambda \mu,
\end{equation}
which leads to:
\begin{equation}
	\lambda < \frac{2}{L \delta^2 \sum_{k=1}^{K} G_k^2}.
\end{equation}
When the above condition is met, the FL algorithm is guaranteed to converge. The learning rate $\lambda$ needs to be inversely proportional to the degree of data heterogeneity (measured by $\delta$). In other words, the more imbalanced the data distribution, the smaller the learning rate $\lambda$ required to ensure convergence.

From the convergence upper bound, it can be observed that the system error mainly consists of artificial noise introduced before the $T_{th}$-th aggregation round and the communication noise interference that persists throughout the entire training process. The level of artificial noise required to meet the DP constraint gradually decreases with the increase in training rounds, while the intensity of communication noise is determined by the transmission power of each device and remains constant during global aggregation.

Higher transmission power reduces the impact of communication noise, which benefits the accuracy of model transmission. However, when the system operates under high transmission power, although the transmission quality is improved, additional artificial noise must be introduced to satisfy the DP constraint. In this case, the convergence error is mainly dominated by the artificial noise. Moreover, the inherent noise in the communication environment fails to effectively contribute to DP satisfaction and instead adds up with the artificial noise, further worsening the error, which is detrimental to system accuracy. Therefore, high transmission power leads to unnecessary resource consumption and degrades system performance, making it a suboptimal strategy.

Reducing transmission power increases the level of communication noise interference. When transmitting the model with excessively low transmission power, although the interference from communication noise may be sufficient to meet DP constraints—requiring little or even no additional artificial noise—the fixed transmission power cannot adapt to the decreasing RoLP. This results in consistently high communication interference, which is also detrimental to global updates. Therefore, the transmission power should not be too low, in order to avoid excessive communication errors that impair system convergence.

In summary, to enhance the convergence performance of FL, it is crucial to fully exploit communication noise as a potential resource for meeting privacy requirements. Specifically, transmission power should neither be too high, which would unnecessarily increase artificial noise, nor too low, which would result in excessive communication errors.
Accordingly, we formulate the optimization problem, as shown in \eqref{problem1}.

\begin{figure*}[t]
    \begin{equation}
    \begin{aligned}
    \mathcal{P}1: \quad  \min_{\boldsymbol{p}} 
		 \frac{1 - A^T}{1 - A} & \sum_{k=1}^{K} \frac{G_k^2 \sigma_{\boldsymbol{n}_0}^2}{\|\boldsymbol{h}_k\|^2 p_k^2}
		+ \sum_{m=0}^{T_{{th}}} A^m \cdot \left( \sum_{k=1}^{K} \frac{2G_k^2 \Delta s_k^2  \ln(\frac{1.25}{\delta_{dp}})}{(\epsilon_k^{[m]})^2} \right), \\
	    s.t.  & \sum_{k=1}^{K} \frac{\sigma_{\boldsymbol{n}_0}}{p_k \|\boldsymbol{h}_k\|} = \sum_{k=1}^{K} \frac{\Delta s_k \sqrt{2 \ln(\frac{1.25}{\delta_{dp}})}}{\epsilon_k^{[T_{{th}}]}},  \\
		 & P_{min} \leq \sum_{k=1}^{K} |p_k|^2 \leq P_{max}. 
    \end{aligned}
    \label{problem1}
    \end{equation}
\hrule
\end{figure*}

Under the privacy constraint, this problem aims to optimize the transmission power allocation of devices to identify the optimal timing $T_{{th}}$ at which the required artificial noise level decreases to match the interference level of the communication environment noise. This allows effective control of total system error, thereby achieving the best trade-off between privacy protection and communication performance.

\subsection{Optimization using DDPG algorithm}

A key challenge in the above optimization problem lies in the fact that the total number of rounds requiring artificial noise, $T_{th}$, dynamically changes with the transmission power allocation decisions. In other words, each optimization attempt of power allocation influences the available privacy budget (i.e., the outcome of LAPA) as well as the convergence upper bound, making the scenario highly coupled and dynamically evolving. Moreover, $T_{th}$ is mathematically difficult to express in a closed form as a function of transmission power $\boldsymbol{p}$. Therefore, traditional static or one-shot offline optimization methods are not effective in handling such strongly coupled and dynamic optimization problems. In contrast, reinforcement learning (RL) is capable of interacting with the environment over multiple rounds, progressively learning an optimal decision policy by repeatedly exploring different power allocation strategies, which enables a more flexible and efficient trade-off among multiple objectives, such as convergence upper bound and privacy noise cost.

Since power allocation is a continuous-variable problem, traditional Deep Q-Network (DQN) algorithms are not well-suited for direct application, as they require discretization of the action space, which leads to accuracy loss and reduced optimization efficiency. To address this, we adopt the Deep Deterministic Policy Gradient (DDPG) algorithm. Through its Actor-Critic structure, DDPG can directly output power allocation decisions in a continuous action space, and employs batch training and target networks to ensure convergence. This enables the system to gradually reach an optimal state that significantly improves convergence performance while also meeting privacy requirements.

We map the $\mathcal{P}1$ problem into a reinforcement learning process, with the following key elements defined:

(1) \textbf{State.} At time step $n$, the state vector is defined as:
\begin{equation}
	\boldsymbol{s}_n = \left[ \boldsymbol{p}^{(n-1)},\ T_{{th}}^{(n)},\ \mathcal{L}^{(n-1)} \right],
\end{equation}
where the elements of the state vector include the power allocation from the previous optimization round, the noise-switching time $T_{{th}}^{(n)}$ determined by the current power allocation, and the objective function value from the previous round. These elements characterize the system’s environment before the current decision.

(2) \textbf{Action.} Since DDPG operates in a continuous action space, the action in the $n$-th round is defined as a new power allocation vector:
\begin{equation}
	\boldsymbol{a}_n = \{ p_k^{(n)} \},\quad k = 1, 2, \dots, K,
\end{equation}
Each action component is a continuous power value, and projection is applied to ensure it satisfies the power constraint $P_{min} \leq \sum_{k=1}^{K} |p_k|^2 \leq P_{max}$.

(3) \textbf{Reward.} The reward is defined as the negative value of the objective function, aiming to unify the learning goal of reward maximization with the objective of minimizing the optimization target:
\begin{equation}
	\boldsymbol{r}_n = -\mathcal{L}\left( \{ p_k^{(n)} \} \right).
\end{equation}
With this design, the agent receives a higher reward when the power allocation results in a lower objective value.

With the DDPG algorithm, we employ an Actor network $\mu(\boldsymbol{s} \mid \boldsymbol{\theta}^\mu)$ and a Critic network $Q(\boldsymbol{s}, a \mid \boldsymbol{\theta}^Q)$ to perform power allocation decision-making and evaluation. Corresponding target networks $\mu'$ and $Q'$ are used for soft updates of the policy and experience replay. Specifically, at each optimization iteration $n$, given the current state $\boldsymbol{s}_n$, the Actor network outputs the power allocation policy as follows:
\begin{equation}
	\boldsymbol{a}_n = \mu(\boldsymbol{s}_n \mid \boldsymbol{\theta}^\mu) + \mathcal{N}_n,
\end{equation}
where $\mathcal{N}_n$ denotes the exploration noise (e.g., Gaussian noise) added to encourage the agent to explore new power configurations. Based on the power allocation vector $\boldsymbol{a}_n$ and the DP-based FL environment defined in this paper, the environment executes one round of FL according to equation (28b) and the dynamic noise control mechanism, determines the noise-switching time $T_{{th}}^{(n)}$, and computes the convergence upper bound $\mathcal{L}^{(n)}$, which yields the immediate reward $\boldsymbol{r}_n = -\mathcal{L}^{(n)}$. The state is then updated to $\boldsymbol{s}_{n+1}$, and the transition tuple $(\boldsymbol{s}_n,\ \boldsymbol{a}_n,\ \boldsymbol{r}_n,\ \boldsymbol{s}_{n+1})$ is stored in the experience replay buffer.

A batch of $N$ data samples $\{(\boldsymbol{s}_n,\ \boldsymbol{a}_n,\ \boldsymbol{r}_n,\ \boldsymbol{s}_{n+1})\}$ is randomly sampled from the experience replay buffer. The target value for each sample is computed using the target networks:
\begin{equation}
	y_i = \boldsymbol{r}_i + \Upsilon Q'\left(\boldsymbol{s}_i',\ \mu'(\boldsymbol{s}_i')\ \middle|\ \boldsymbol{\theta}^{Q'}\right),
\end{equation}
where $\Upsilon$ is the discount factor. The Critic network parameters $\boldsymbol{\theta}^Q$ are then updated by performing gradient descent to minimize the mean squared error loss:
\begin{equation}
	\mathcal{L}(\theta^Q) = \frac{1}{N} \sum_{i=1}^{N} \left( y_i - Q(\boldsymbol{s}_i,\ \boldsymbol{a}_i\ \middle|\ \boldsymbol{\theta}^Q) \right)^2.
\end{equation}

The Actor network is updated using the deterministic policy gradient to output better actions that maximize the Q-values estimated by the Critic network:
\begin{equation}
\begin{aligned}
    	&\nabla_{\boldsymbol{\theta}^\mu} J \approx \frac{1}{N} \\
        & \cdot \sum_{i=1}^{N} \left[ 
    	\left. \nabla_a Q(\boldsymbol{s},\boldsymbol{a}\ \middle|\ \boldsymbol{\theta}^Q) \right|_{\boldsymbol{s}=\boldsymbol{s}_i,\ \boldsymbol{a}=\mu(\boldsymbol{s}_i\ \mid\ \boldsymbol{\theta}^\mu)} 
    	 \nabla_{\boldsymbol{\theta}^\mu} \mu(\boldsymbol{s}_i\ \mid\ \boldsymbol{\theta}^\mu) \right].
\end{aligned}
\end{equation}

Finally, soft updates are used to smoothly update the target network parameters:
\begin{equation}
	\boldsymbol{\theta}^{Q'} \leftarrow \tau \boldsymbol{\theta}^Q + (1 - \tau)\boldsymbol{\theta}^{Q'},\quad
	\boldsymbol{\theta}^{\mu'} \leftarrow \tau \boldsymbol{\theta}^\mu + (1 - \tau)\boldsymbol{\theta}^{\mu'}.
\end{equation}
where $\tau \ll 1$, typically set to $0.001$.

The above steps are repeated until the maximum number of training episodes is reached. Through this process, this section achieves an effective solution to the continuous power allocation optimization problem using the DDPG algorithm. 

\subsection{Complexity Analysis}

The computational complexity of the DDPG algorithm adopted in this paper mainly arises from two aspects: neural network and environment simulation. Specifically, in each iteration, the algorithm performs a forward pass from state to action with a complexity of $\mathcal{O}(W)$, where $W$ denotes the total number of neural network parameters. The environment simulation involves determining the noise-switching time $T_{{th}}$ under a given power allocation and executing the corresponding FL process and DP mechanism, with a complexity denoted by $\mathcal{O}(\alpha \cdot T_{{th}})$. In addition, each training step samples a batch of $N$ data points from the replay buffer to update the parameters of the Critic and Actor networks, with a complexity of approximately $\mathcal{O}(N \cdot W)$. Therefore, the overall complexity per iteration is:
\begin{equation}
	\mathcal{O}(W + N \cdot W + \alpha \cdot T_{{th}}),
\end{equation}
Considering that the algorithm typically runs for $E \cdot T_{{RL}}$ iterations (i.e., $E$ episodes with $T_{{RL}}$ steps per episode), the total complexity is:
\begin{equation}
	\mathcal{O}\left(E \cdot T_{{RL}} \cdot (W + N \cdot W + \alpha \cdot T_{{th}})\right).
\end{equation}

\section{Simulation Results}

\subsection{Simulation Configuration}

\begin{figure}
	\centerline{\includegraphics[width=1\linewidth]{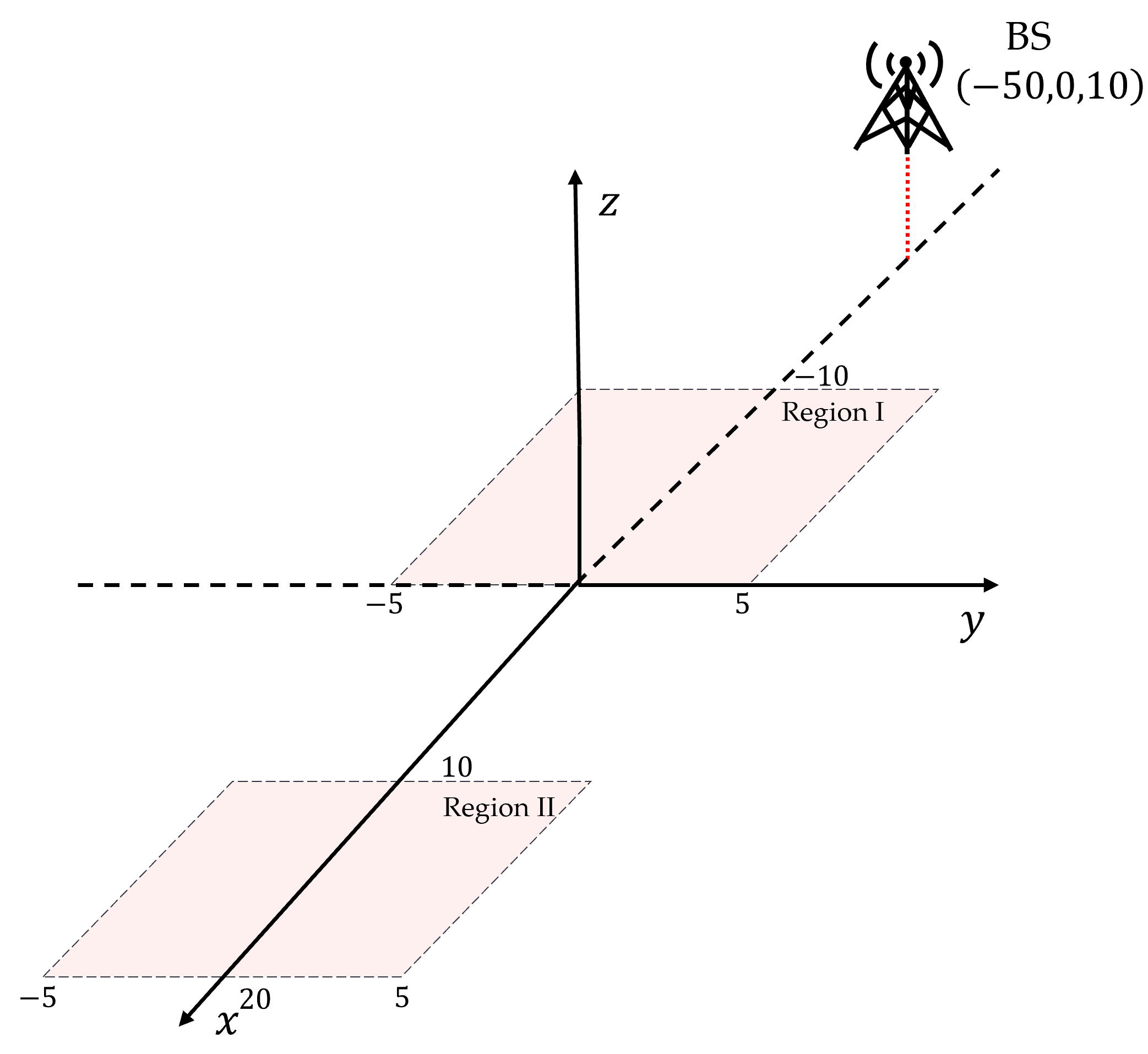}}
	\caption{Device distribution in the simulation environment.}
	\label{simulationlayout}
\end{figure}

We adopt a Cartesian coordinate system to describe the spatial relationships among devices in the simulation, as shown in Fig.~\ref{simulationlayout}. The BS, acting as the parameter server, is located at $(-50,\ 0,\ 10)\ \text{m}$. A total of $K = 15$ devices are distributed across two regions. Specifically, 7 devices are randomly selected to reside in Region I, defined as $\{(x,\ y,\ 0):\ -10 \leq x \leq 0,\ -5 \leq y \leq 5\}\ \text{m}$, while the remaining 8 devices are placed in Region II, defined as $\{(x,\ y,\ 0):\ 10 \leq x \leq 20,\ -5 \leq y \leq 5\}\ \text{m}$. Different Non-IID data partitions significantly affect the experimental results. In this experiment, lower label diversity reduces the randomness of the data, making the training process more stable and efficient. We employ stochastic gradient descent (SGD) for local model training, with a batch size of $0.1$. The learning rate is set to $\lambda = 0.01$ for the MNIST dataset and $\lambda = 0.008$ for the Fashion-MNIST dataset.

The path loss in the wireless transmission system is given by:
\begin{equation}
	{PL}_{{DB}} = G_{{BS}} G_{{D}} \left( \frac{c}{4\pi f_c d_{{DB}}} \right)^P,
\end{equation}
where the number of antennas at the BS is $N_a = 15$, the antenna gain at the BS is $G_{{BS}} = 5\ \text{dBi}$, and the device antenna gain is $G_{{D}} = 0\ \text{dBi}$. The carrier frequency is $f_c = 915\ \text{MHz}$, the path loss exponent is $P = 3.76$, and $d_{\text{DB}}$ denotes the distance between the device and the BS. The speed of light is denoted by $c$, and the noise power is set to $10^{-3}\ \text{W}$.

We use a CNN network to train and test the MNIST and Fashion-MNIST datasets, which consists of two $5 \times 5$ convolutional layers (each followed by a $2 \times 2$ max pooling layer), followed by a batch normalization layer, a fully connected layer with 50 units, a ReLU activation layer, and a softmax output layer. The cross-entropy function is used as the loss function during training.

Devices are distributed in different regions, and the data samples are clearly imbalanced across devices. Due to channel heterogeneity (mainly caused by varying distances) and data heterogeneity, the straggler problem may become more severe. Under such conditions, the proposed LAPA-based dynamic noise control strategy and customized aggregation weights are particularly important for improving learning performance under privacy constraints. This will be verified through the following simulation experiments.

\subsection{Non-IID Aggregation with Device Selection Mechanism}

We evaluate the effectiveness and superiority of the personalized aggregation strategy proposed in Section III-C under heterogeneous scenarios using accuracy on both datasets. The simulation is conducted under practical communication conditions. To assess the impact of different SINR thresholds used in the device selection mechanism on convergence and accuracy, we convert the SINR threshold in ~\eqref{6 G_k} into an SER threshold $\epsilon_{{SER}}$. The following Configurations are defined for simulation:

\textbf{Configuration 1:} The aggregation strategy proposed in this paper, without device selection;

\textbf{Configuration 2:} The proposed aggregation strategy with device selection based on SER performance, using different filtering thresholds: $10^{-1}$, $10^{-2}$, and $10^{-3}$.

\begin{figure}
    \begin{subfigure}{\columnwidth} 
        \centering
        \includegraphics[width=0.98\columnwidth]{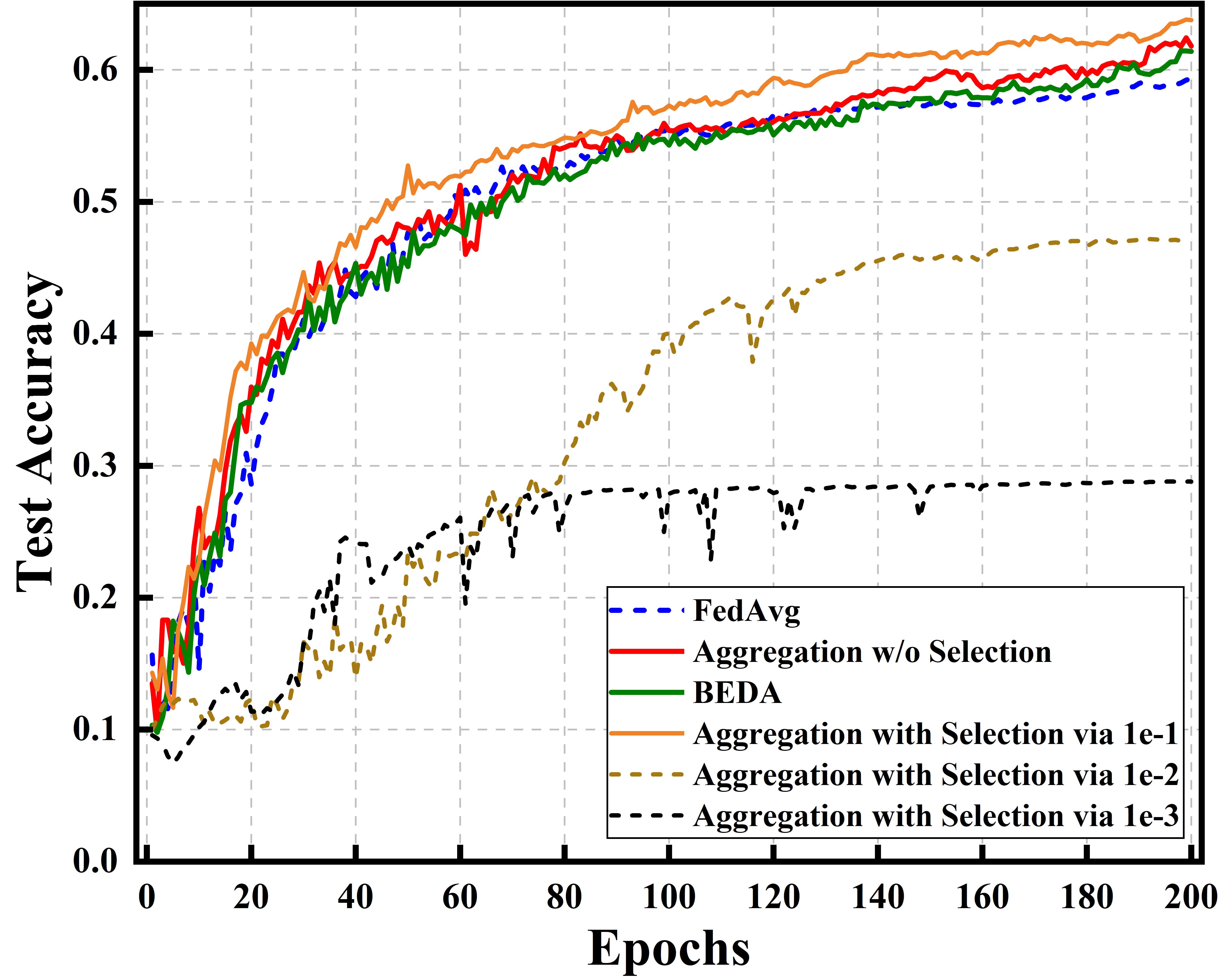} 
        \caption{MNIST}
        \includegraphics[width=0.98\columnwidth]{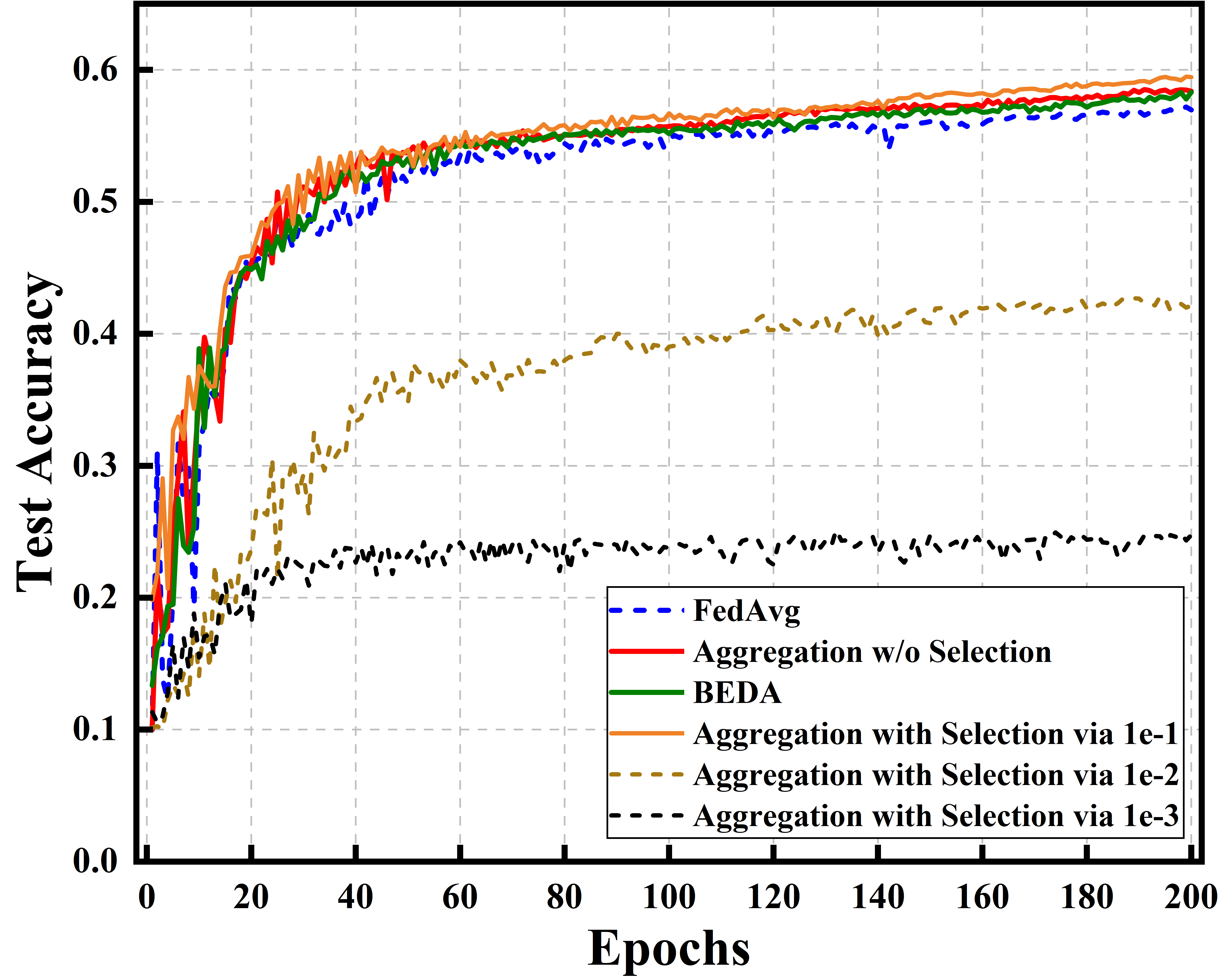} 
        \caption{Fashion-MNIST}
    \end{subfigure}
    \caption{Comparison of different aggregation strategies in a heterogeneous wireless communication environment.}
\label{exp1}
\end{figure}

In addition, we compare the following approaches: (1) the traditional Federated Averaging aggregation strategy (FedAvg), referred to as \textbf{Benchmark 1}; and (2) a widely used aggregation strategy designed for heterogeneous data distributions, referred to as \textbf{Benchmark 2}, namely the Benefit Evaluation-based Dynamic Aggregation (BEDA) algorithm~\cite{wu2021fast}, which constructs aggregation weights based on the angle between local and global gradients. Although this method can quantify contribution in real time, it suffers from instability when data distributions are highly imbalanced, as the local gradient variability directly affects the reliability of contribution estimation. The simulation results are shown in Fig.~\ref{exp1}.

First, it can be observed that under Non-IID data distribution, both the proposed aggregation strategy based on Wasserstein distance (Configuration 1) and the angle-based aggregation strategy in Benchmark 2 significantly improve FL accuracy compared to the traditional FedAvg algorithm. Specifically, on the MNIST dataset, Configuration 1 improves accuracy by 2.470\% over Benchmark 1, while Benchmark 2 improves it by 2.073\%. On the Fashion-MNIST dataset, Configuration 1 improves by 1.413\% and Benchmark 2 by 1.333\%. This indicates that under Non-IID data distribution, it is necessary to design customized aggregation weights tailored to the data distribution of different devices to mitigate the update bias caused by data imbalance.

Second, under practical communication conditions, the proposed Wasserstein distance-based aggregation strategy outperforms the widely used Benchmark 2 in terms of aggregation performance. Specifically, Configuration 1 achieves improvements of 0.397\% (MNIST) and 0.800\% (Fashion-MNIST) over Benchmark 2. Furthermore, by incorporating a device selection mechanism based on transmission quality to exclude devices with severe transmission errors, accuracy can be further improved. When the SER threshold is set to $\epsilon_{{SER}} = 10^{-1}$, the performance is optimal, with improvements of 2.367\% (MNIST) and 1.137\% (Fashion-MNIST) over Benchmark 2. This is because devices with severe transmission errors are harmful to the FL update. However, setting a more stringent selection threshold that excludes more devices with even slight errors leads to reduced accuracy, since in Non-IID FL, having more valid data is beneficial to model updates.

In summary, under Non-IID conditions, it is important to assign customized and stable aggregation weights based on the data distribution of devices. Moreover, in noisy environments, devices with Non-IID data should provide sufficient and accurate aggregated data for the learning system. Therefore, the proposed aggregation strategy based on Wasserstein distance with device selection capability demonstrates clear advantages.

\subsection{Superiority of the LAPA Algorithm}

This simulation is designed to conduct a horizontal comparison of the performance of different privacy budget allocation strategies applied in the FL system. The simulation is carried out under ideal communication conditions. The Configurations are defined as follows:

\textbf{Configuration 3:} The LAPA algorithm proposed in this paper;

\textbf{Configuration 4:} No privacy allocation mechanism is applied in the FL process.

In addition, we compare with the traditional DP algorithm based on uniform artificial noise~\cite{abadi2016deep}, referred to as \textbf{Benchmark 3}, and an advanced existing method, the Adaptive Rayleigh Budgeting (ARB) algorithm~\cite{zhang2021privacy}, referred to as \textbf{Benchmark 4}. The ARB algorithm assigns personalized privacy budgets to devices based on global gradient updates, but does not take into account the local update bias caused by Non-IID data.

\begin{figure}
	\centerline{\includegraphics[width=1\linewidth]{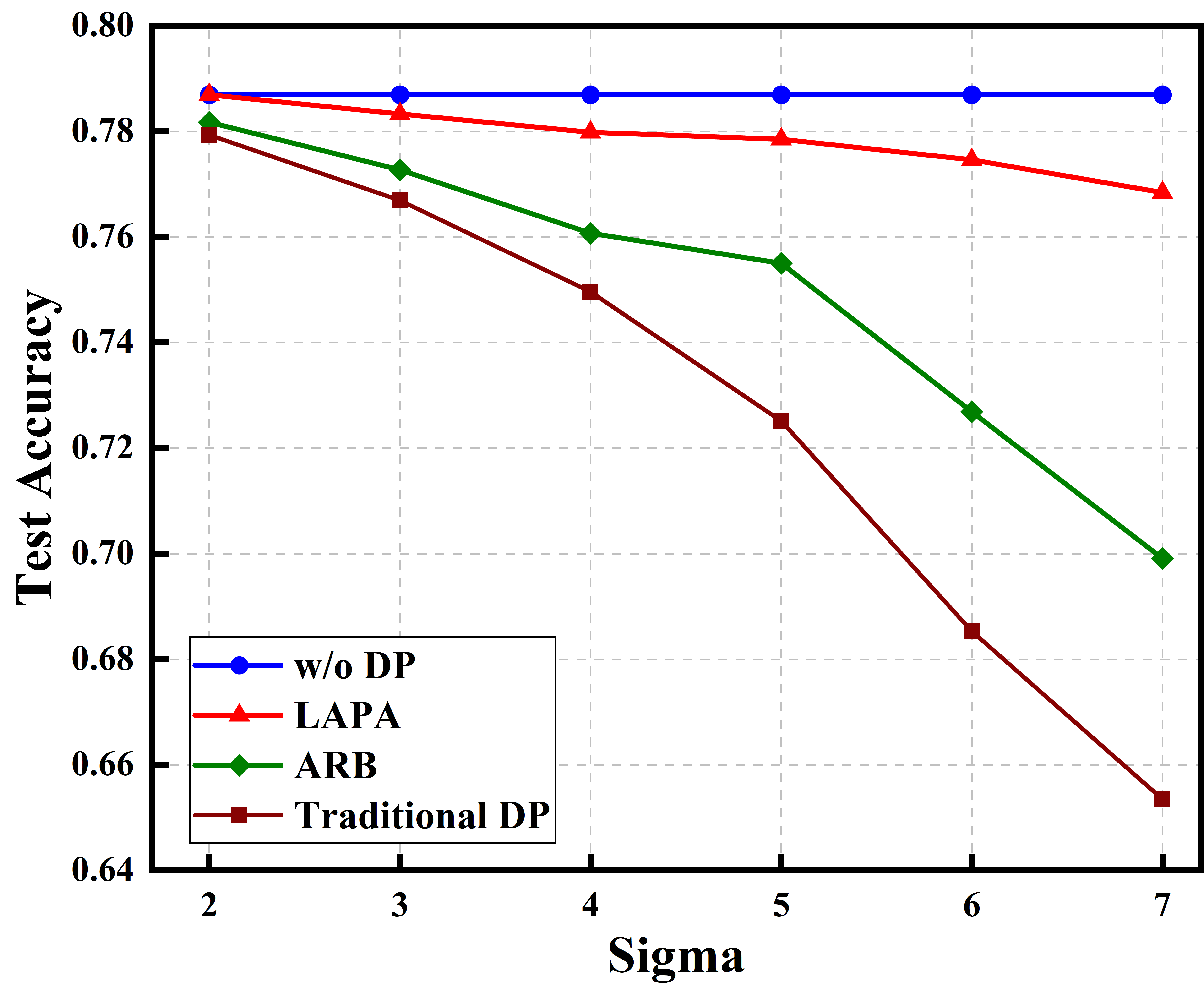}}
	\caption{FL performance of different privacy allocation strategies under various artificial noise levels.}
	\label{exp2_1}
\end{figure}

We first compare the FL performance of each scheme under different initial noise levels on the MNIST dataset, as shown in Fig.~\ref{exp2_1}. The vertical axis represents the final convergence accuracy of FL when applying different privacy allocation strategies, while the horizontal axis indicates the initial level of artificial noise for each strategy (unit: $W$).

As shown in the figure, omitting a privacy strategy in FL maintains the most stable and optimal aggregation performance. However, such a strategy does not provide the FL system with any resistance against differential attacks, and thus only serves as a performance reference for other DP-based strategies. The three types of DP strategies exhibit different performance fluctuations under varying levels of artificial noise. This indicates that under the same privacy level, different DP strategies result in varying degrees of degradation in learning performance.
Under Non-IID condition, both the ARB and LAPA strategies demonstrate clear advantages in aggregation performance compared to the traditional DP strategy. Notably, the LAPA algorithm achieves convergence performance nearly equivalent to that of FL without any DP noise, suggesting that it can provide personalized privacy budget allocation based on heterogeneous data distributions while still meeting DP requirements, preserving FL convergence performance to the greatest extent.

\begin{figure*}[htbp]
    \centering
    \begin{minipage}{0.18\linewidth}
        \centering
        \quad \small{\small{3 IID+\textcolor{red}{12 Non-IID}}}
    \end{minipage}
    \begin{minipage}{0.18\linewidth}
        \centering
        \quad \small{\small{5 IID+\textcolor{red}{10 Non-IID}}}
    \end{minipage}
    \begin{minipage}{0.18\linewidth}
        \centering
        \quad \small{\small{7 IID+\textcolor{red}{8 Non-IID}}}
    \end{minipage}
    \begin{minipage}{0.18\linewidth}
        \centering
        \quad \small{\small{9 IID+\textcolor{red}{6 Non-IID}}}
    \end{minipage}
    \begin{minipage}{0.18\linewidth}
        \centering
        \quad \small{\small{0 IID+\textcolor{red}{15 Non-IID}}}
    \end{minipage}\\[3pt]
    
    \begin{minipage}{0.18\linewidth}
        \centering
        \includegraphics[width=\linewidth]{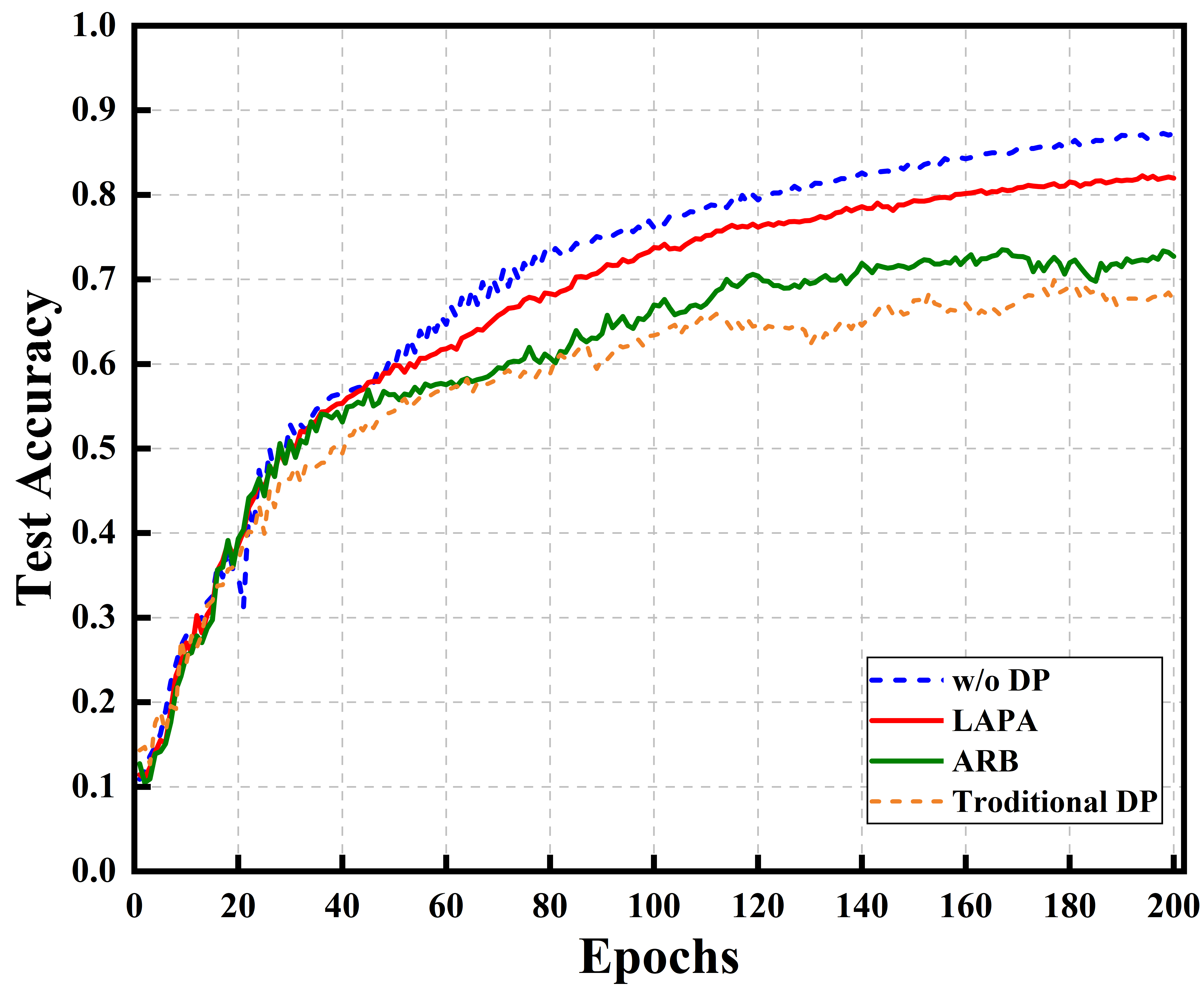}
    \end{minipage}
    \begin{minipage}{0.18\linewidth}
        \centering
        \includegraphics[width=\linewidth]{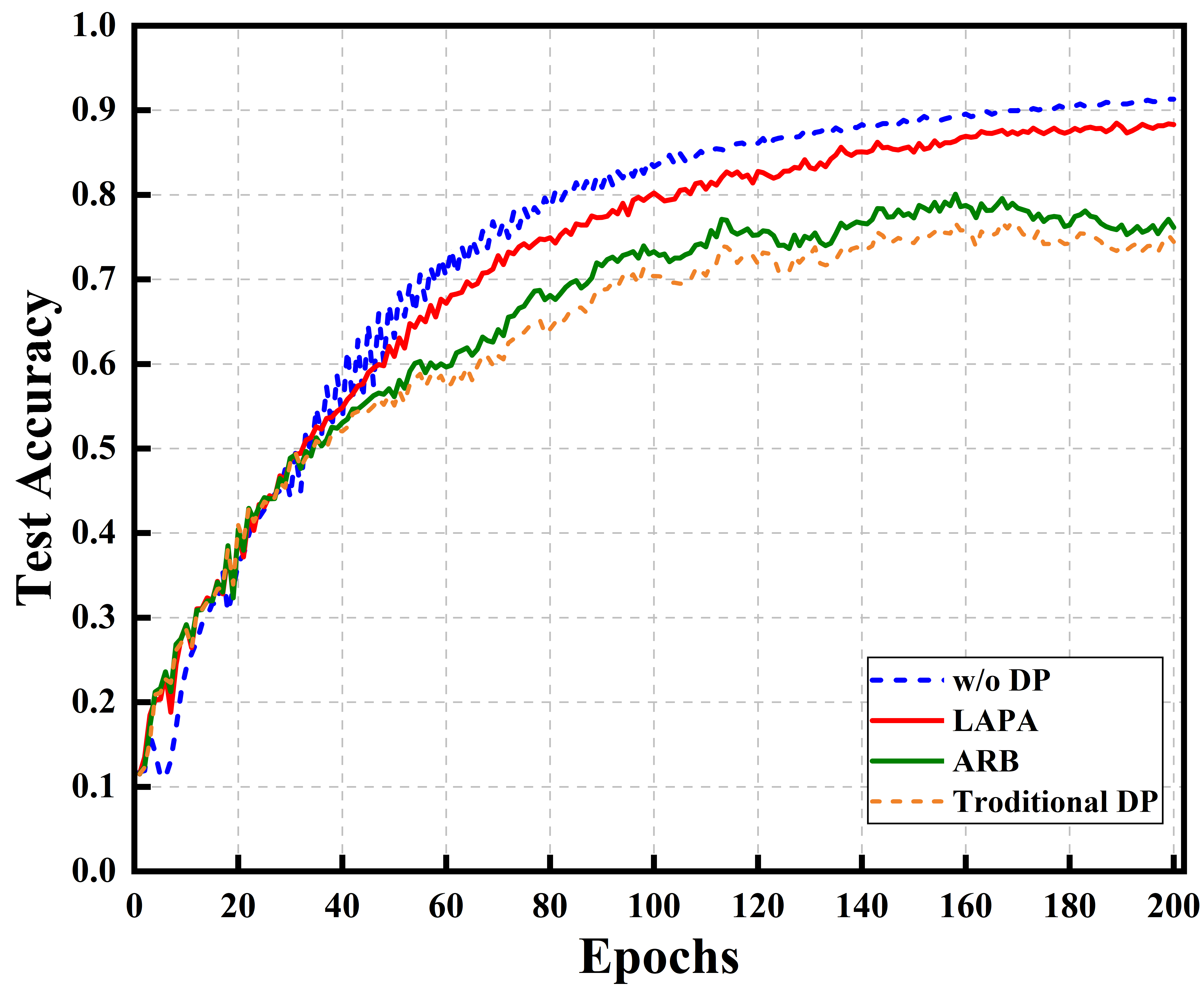}
    \end{minipage}
    \begin{minipage}{0.18\linewidth}
        \centering
        \includegraphics[width=\linewidth]{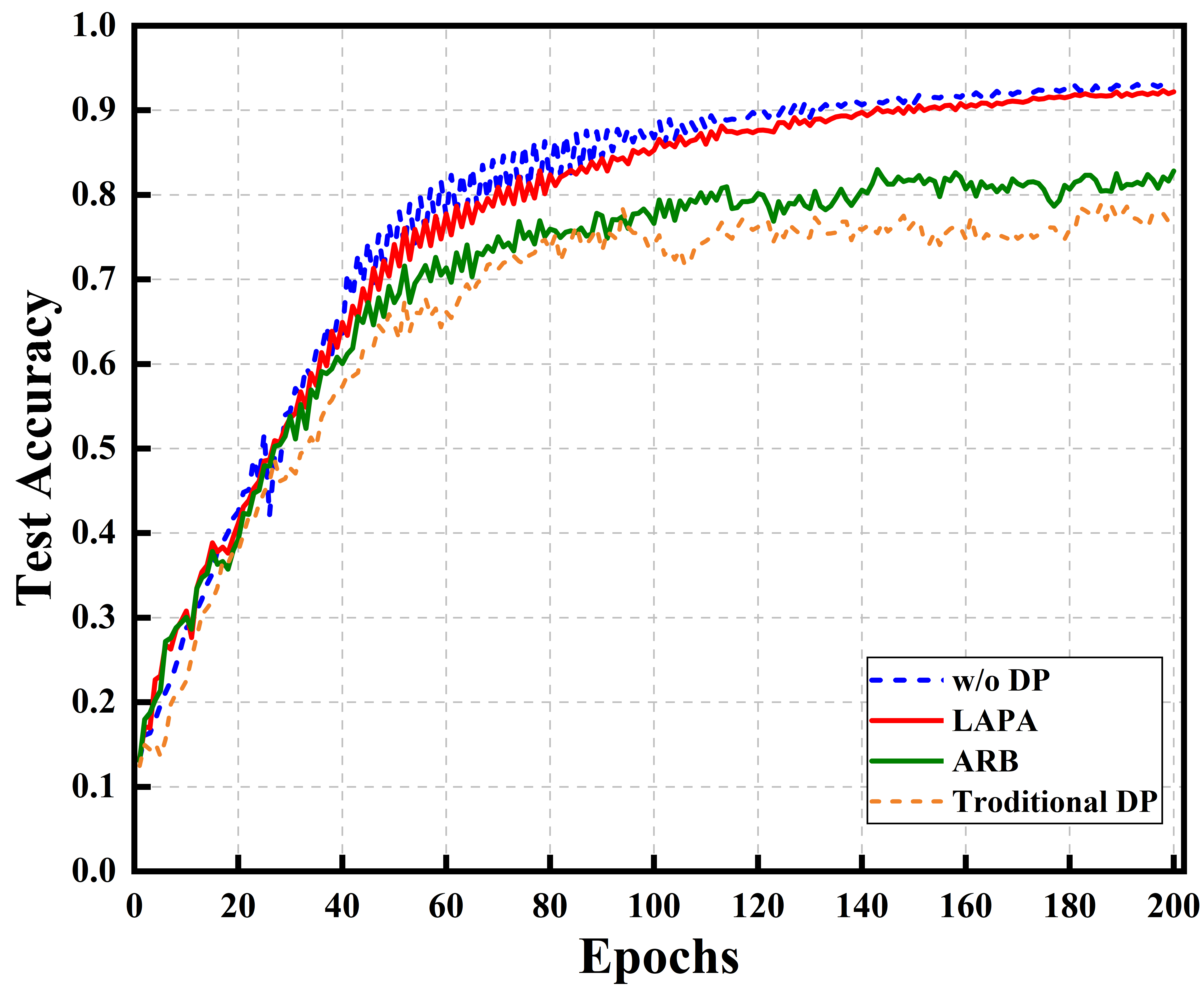}
    \end{minipage}
    \begin{minipage}{0.18\linewidth}
        \centering
        \includegraphics[width=\linewidth]{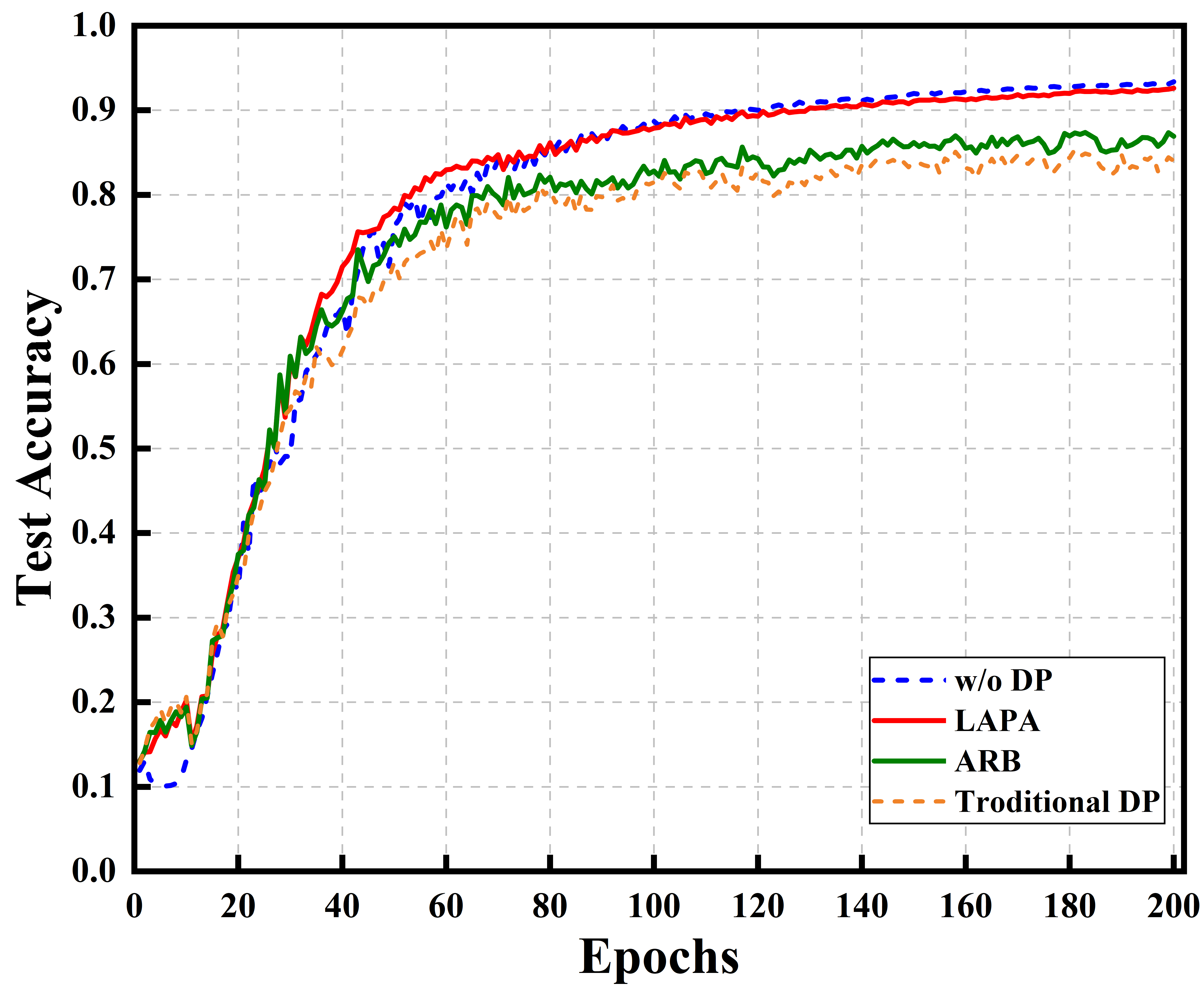}
    \end{minipage}
    \begin{minipage}{0.18\linewidth}
        \centering
        \includegraphics[width=\linewidth]{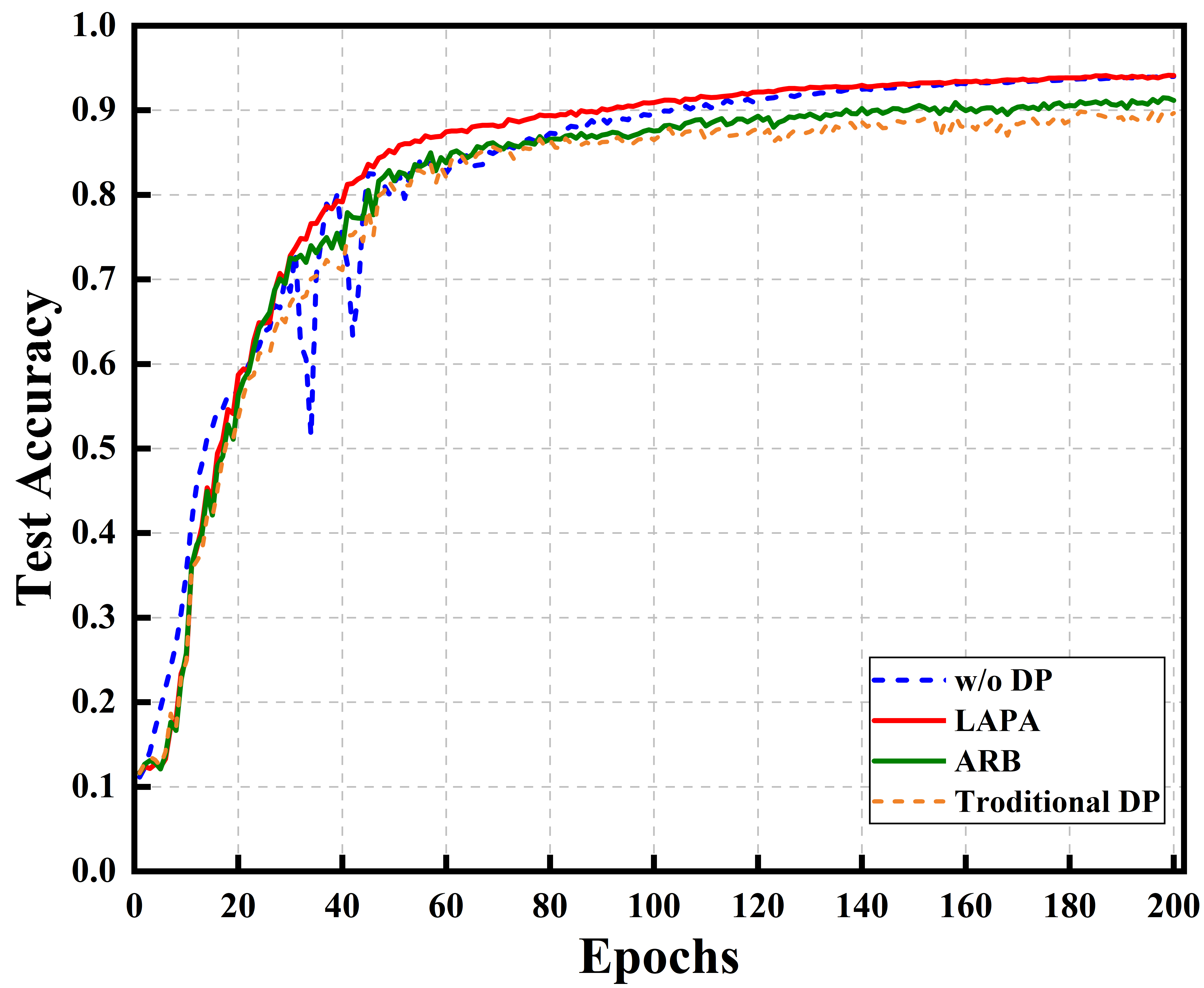}
    \end{minipage}
    \\[5pt] \parbox{0.5\linewidth}{\centering (a) MNIST}\\[5pt]

    \begin{minipage}{0.18\linewidth}
        \centering
        \includegraphics[width=\linewidth]{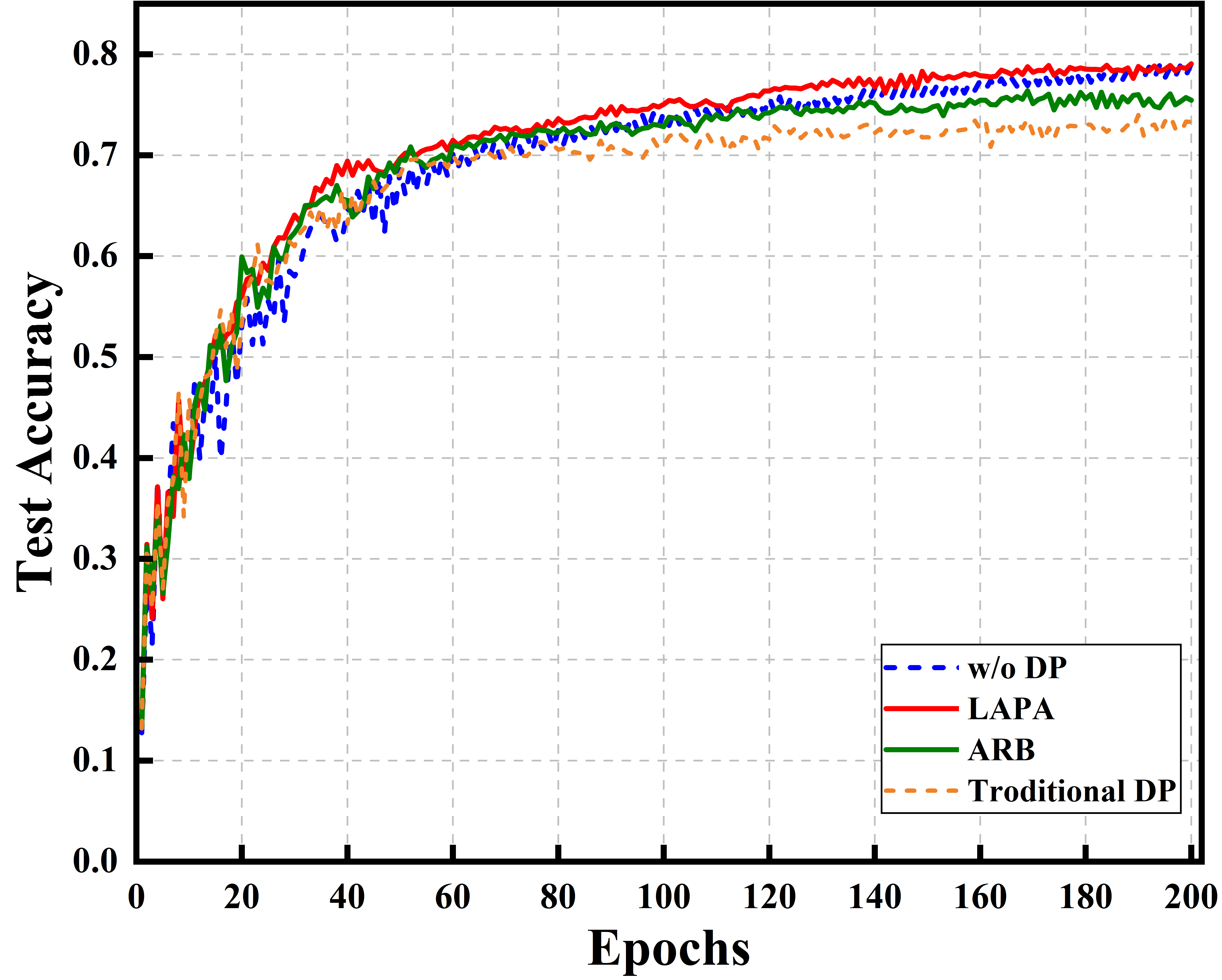}
    \end{minipage}
    \begin{minipage}{0.18\linewidth}
        \centering
        \includegraphics[width=\linewidth]{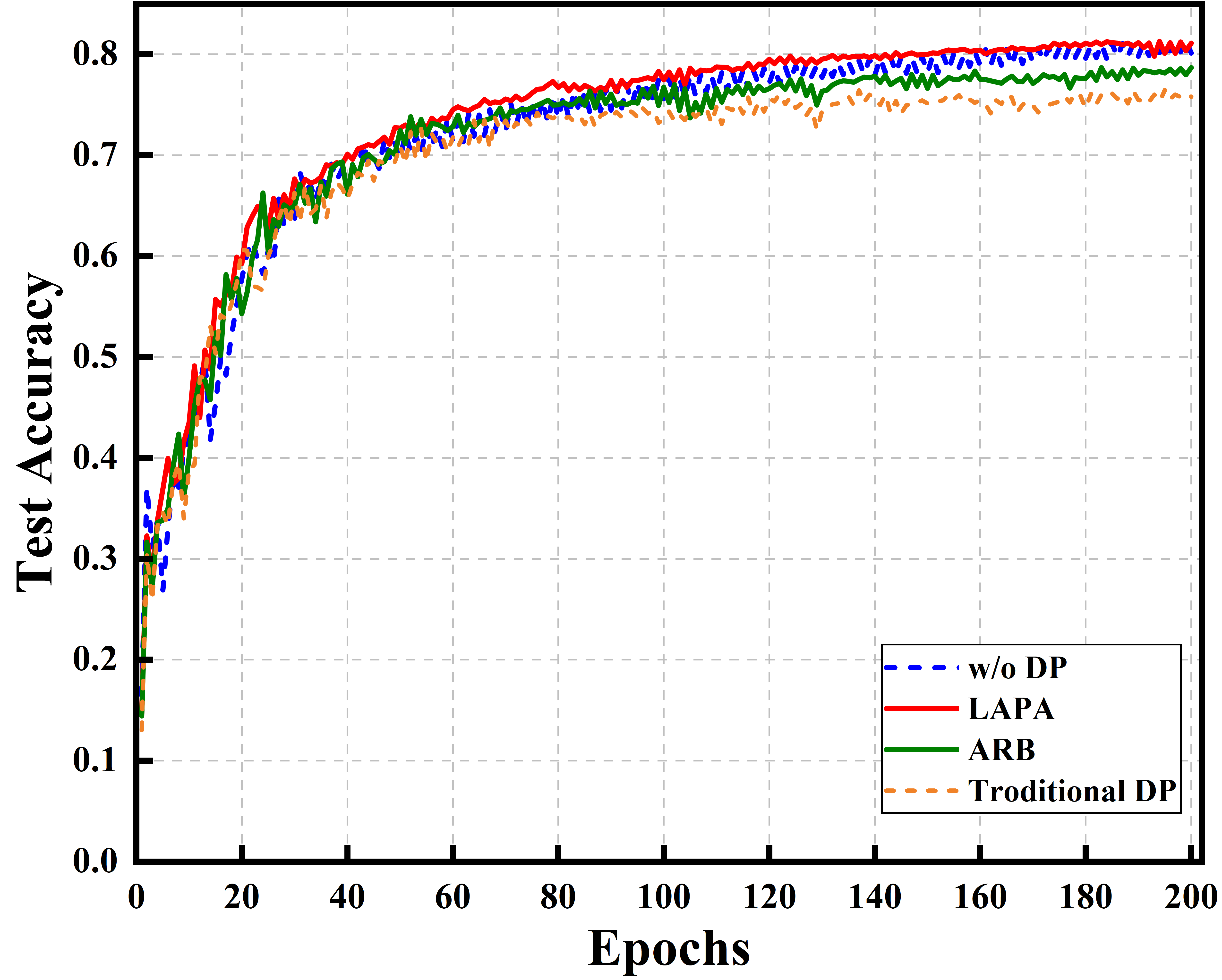}
    \end{minipage}
    \begin{minipage}{0.18\linewidth}
        \centering
        \includegraphics[width=\linewidth]{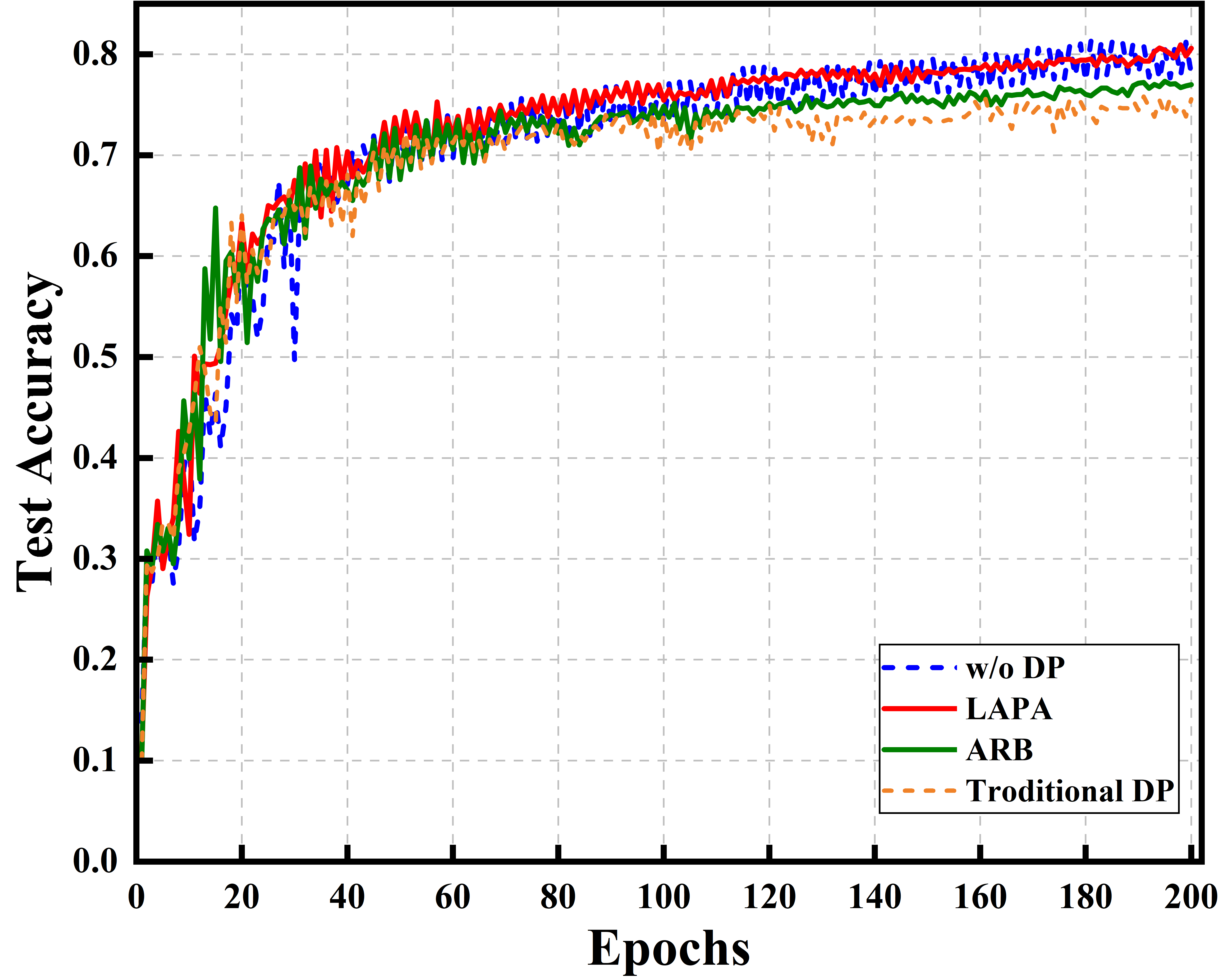}
    \end{minipage}
    \begin{minipage}{0.18\linewidth}
        \centering
        \includegraphics[width=\linewidth]{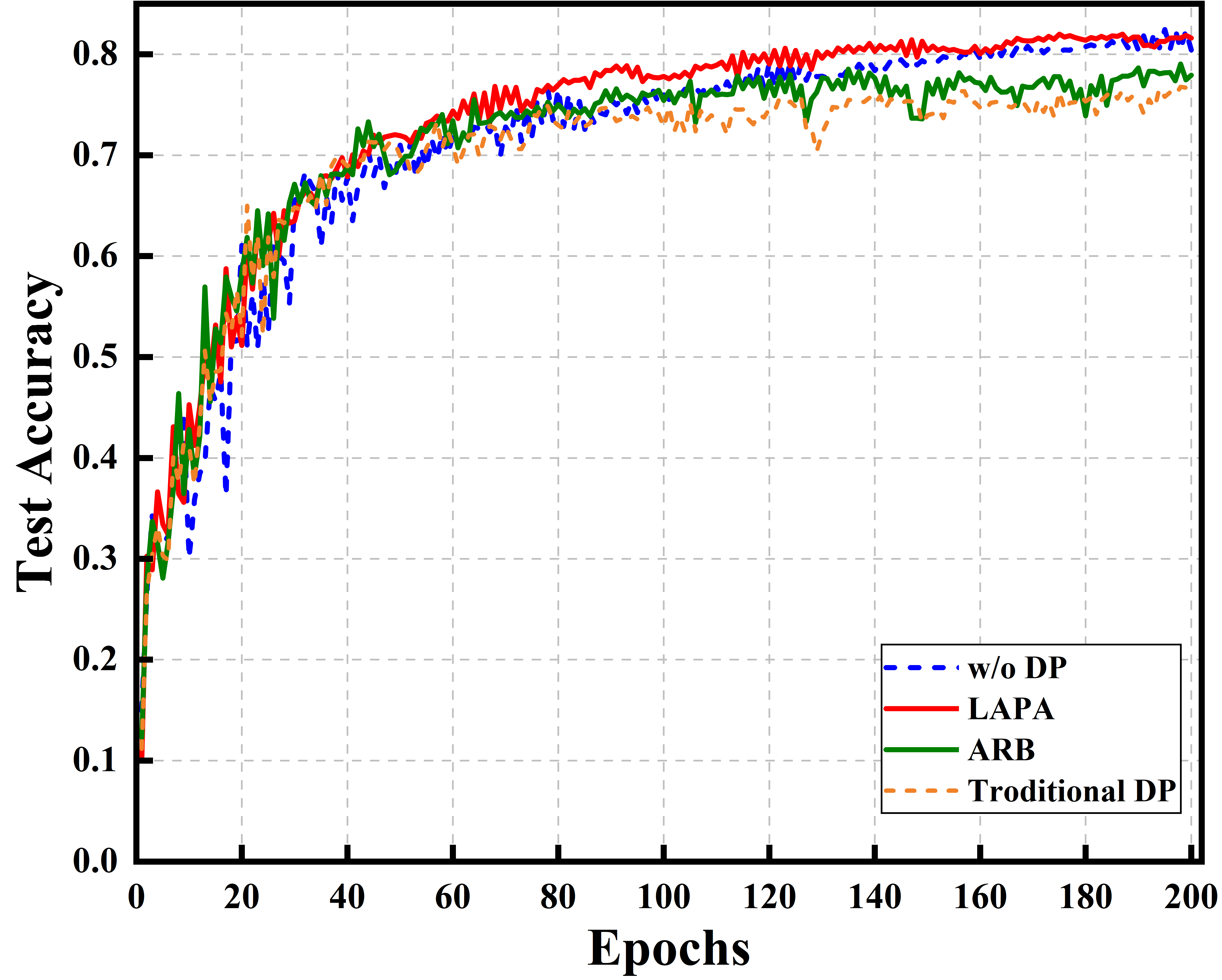}
    \end{minipage}
    \begin{minipage}{0.18\linewidth}
        \centering
        \includegraphics[width=\linewidth]{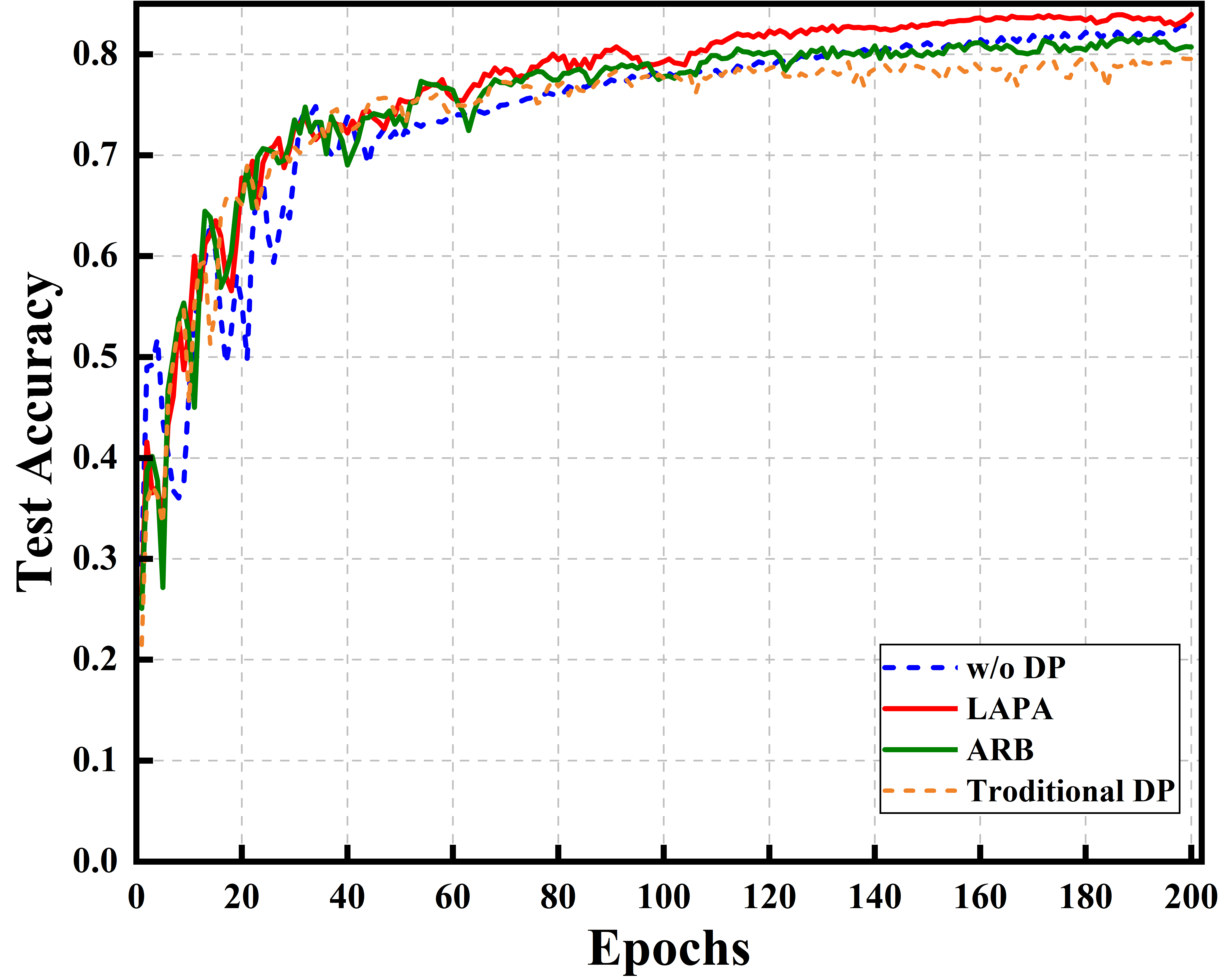}
    \end{minipage}
     \\[5pt] \parbox{0.5\linewidth}{\centering (b) Fashion-MNIST}

    \caption{FL convergence performance of different privacy allocation strategies under varying degrees of data heterogeneity.}
    \label{exp2_2}
\end{figure*}

Next, we assign different data distributions to devices and observe the FL convergence performance under varying degrees of data heterogeneity for each privacy strategy, as shown in Fig.~\ref{exp2_2}.

As shown in the figure, regardless of the degree of data heterogeneity, the LAPA algorithm proposed in this paper consistently outperforms Benchmark 3 and Benchmark 4. Additionally, we observe an interesting phenomenon: when the learning rate is fixed, as the proportion of IID devices increases, the performance of the LAPA-based strategy gradually approaches that of the non-DP strategy. When all device data distributions are IID, the convergence rate under LAPA even slightly exceeds that of the non-DP case (although with continued training, the final accuracy remains slightly lower than the non-DP case). This phenomenon arises because, under well data conditions, the system typically requires a larger learning rate. However, when a smaller learning rate is used for stability, the LAPA algorithm introduces just enough artificial noise, which in turn helps alleviate underfitting. As a result, the introduced artificial noise not only minimizes the impact on model convergence while ensuring privacy, but also mitigates underfitting in certain ideal scenarios (thanks to the controlled noise intensity). Furthermore, from a reverse perspective, as the proportion of IID devices decreases, the data-adaptive privacy mechanism demonstrates stronger adaptability, showing good robustness against variations in data heterogeneity.

\subsection{Dynamic Noise Control Optimization Based on the LAPA Algorithm}

\begin{figure}
    \begin{subfigure}{\columnwidth} 
        \centering
        \includegraphics[width=0.98\columnwidth]{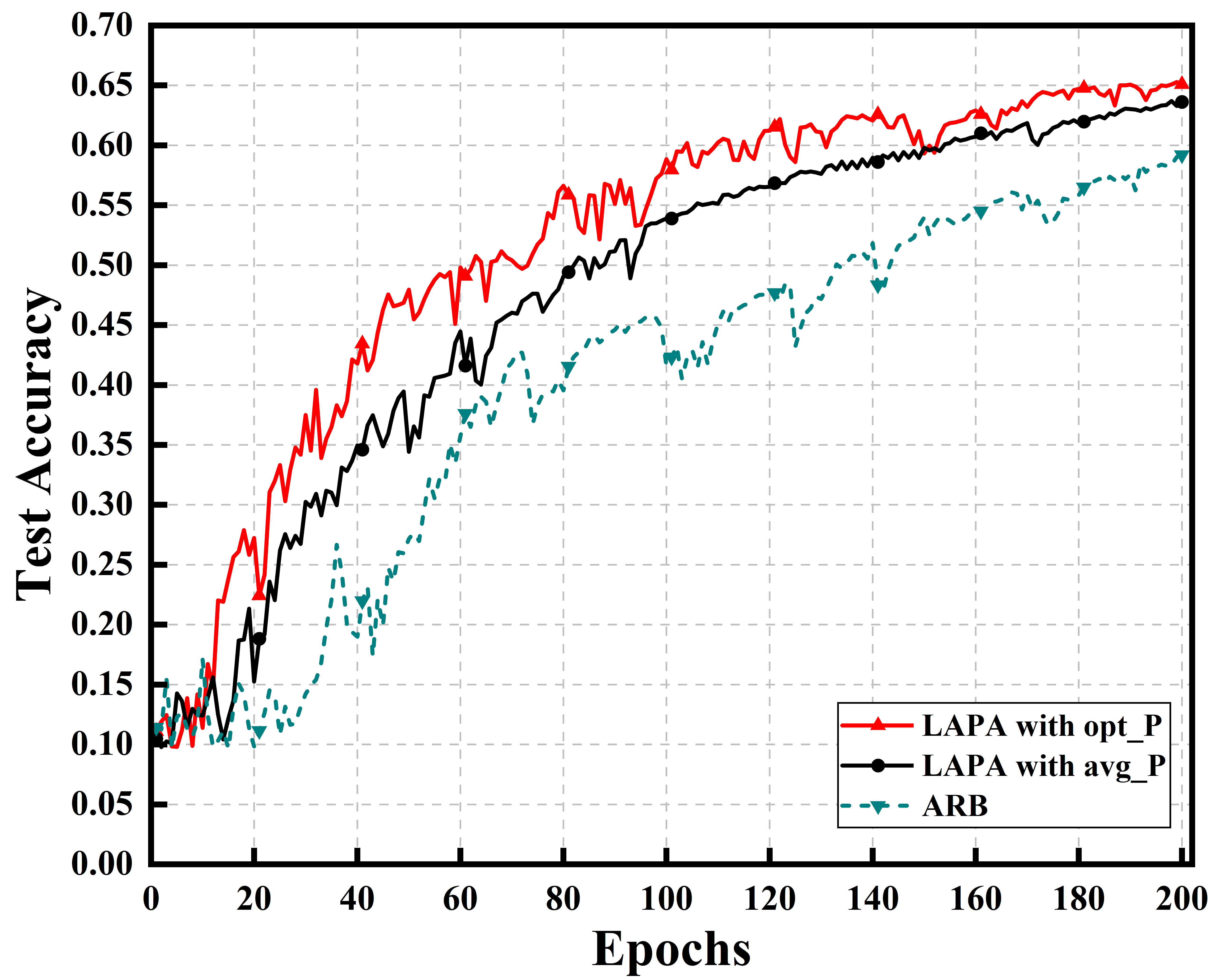} 
        \caption{MNIST}
        \includegraphics[width=0.98\columnwidth]{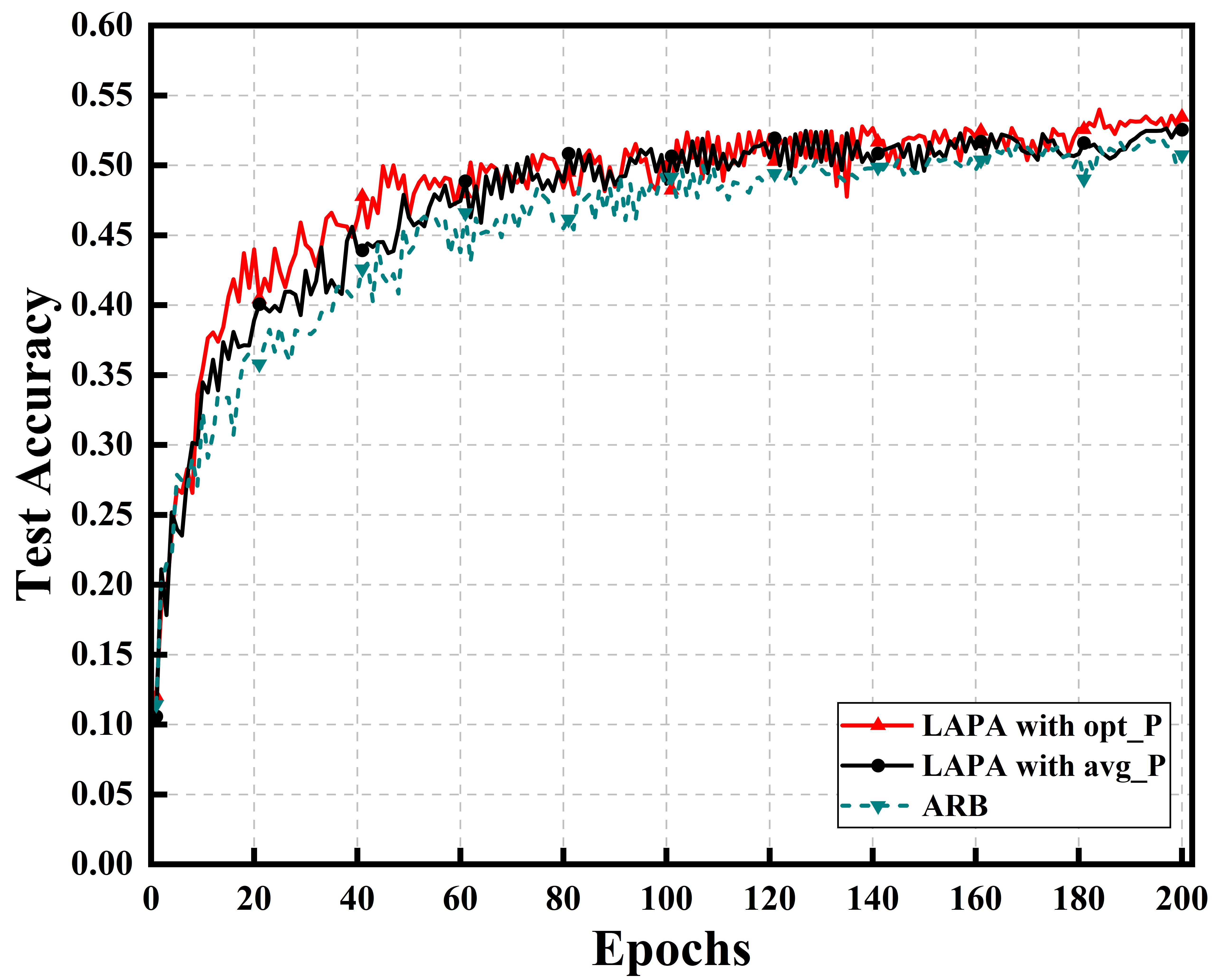} 
        \caption{Fashion-MNIST}
    \end{subfigure}
    \caption{Superiority of the proposed dynamic noise control optimization strategy based on the LAPA algorithm}
\label{exp3}
\end{figure}

As shown in the simulation results of the previous section, under ideal communication conditions and Non-IID data distribution, both the proposed LAPA algorithm and the advanced ARB algorithm can improve FL performance under privacy protection compared to the traditional DP algorithm, with the LAPA algorithm outperforming ARB. In this simulation, we evaluate the superiority of the LAPA-based dynamic noise control optimization strategy (denoted as \textbf{Configuration 5}) under realistic communication noise conditions. The simulation results are presented in Fig.~\ref{exp3}.

As shown in Fig.~\ref{exp3}, on the MNIST dataset, Configuration 5 achieves a 1.465\% improvement compared to its average power setting, and a 5.872\% improvement compared to the average power setting of Benchmark 4. On the Fashion-MNIST, the improvements are 0.910\% and 2.705\%, respectively. This demonstrates that, under realistic noise conditions, optimizing transmission power to achieve an optimal balance between artificial and channel noise can effectively leverage environmental noise interference, maximizing system utility while ensuring DP guarantees.

\section{Conclusion}
This paper addresses the problem of privacy protection and aggregation performance optimization in FL under Non-IID data condition by proposing a lightweight adaptive privacy allocation (LAPA) strategy. This strategy dynamically adjusts the privacy budget based on gradient contribution without introducing additional communication overhead, and utilizes the inherent noise in the communication environment to optimize the timing of artificial noise injection, aiming to balance privacy protection and system utility. In addition, based on data distribution characteristics and channel quality, an efficient device selection and aggregation weight assignment mechanism is designed, which improves convergence performance while ensuring privacy. Theoretical analysis and simulation results validate the effectiveness of the proposed method.


\section{Appendix I}

In this paper, the total noise introduced into the FL system is denoted as $N_k^t$, which includes the artificial noise added before the $T_{{th}}$-th iteration, as well as the channel noise that persists throughout the process. Thus, we have:
\begin{equation}
\begin{aligned}
    	&\boldsymbol{w}^{[t+1]} - \boldsymbol{w}^{[t]} \\
    	&= -\lambda \cdot \sum_{k=1}^{K} G_k \left[ \nabla F_k(\boldsymbol{w}^{[t]}) + N_k^t \right] \\
    	&= -\lambda \cdot \left[ \sum_{k=1}^{K} G_k \nabla F_k(\boldsymbol{w}^{[t]}) + \sum_{k=1}^{K} G_k N_k^t \right].
\end{aligned}
\end{equation}
Therefore,
\begin{equation}
	\begin{split}
		&\mathbb{E}\left[\left\| \boldsymbol{w}^{[t+1]} - \boldsymbol{w}^{[t]} \right\|^2\right] \\
        &= \lambda^2 \cdot \mathbb{E}\left[\left\| \sum_{k=1}^{K} G_k \nabla F_k(\boldsymbol{w}^{[t]}) + \sum_{k=1}^{K} G_k N_k^t \right\|^2\right].
	\end{split}
\end{equation}

Since we assumes that the artificial noise added before the $T_{{th}}$-th iteration follows a zero-mean Gaussian distribution, we have:
\begin{equation}
\begin{aligned}
    	&\mathbb{E}\left[\left\| \boldsymbol{w}^{[t+1]} - \boldsymbol{w}^{[t]} \right\|^2\right] \\
        &= \lambda^2 \left\{ \mathbb{E}\left[\left\| \sum_{k=1}^{K} G_k \nabla F_k(\boldsymbol{w}^{[t]}) \right\|^2\right] + \mathbb{E}\left[\left\| \sum_{k=1}^{K} G_k N_k^t \right\|^2\right] \right\}.
\end{aligned}
\end{equation}

By performing a Taylor expansion of $F(\boldsymbol{w}^{[t+1]})$ and taking the expectation, we obtain:
\begin{equation}
	\begin{aligned}
		&\mathbb{E}[F(\boldsymbol{w}^{[t+1]})] \\
		&\leq \mathbb{E}[F(\boldsymbol{w}^{[t]})] - \lambda \left\| \nabla F(\boldsymbol{w}^{[t]}) \right\|^2  \\
		& \quad +\frac{L\lambda^2}{2} \left\{ \mathbb{E}\left[\left\| \sum_{k=1}^{K} G_k \nabla F_k(\boldsymbol{w}^{[t]}) \right\|^2 \right] + \mathbb{E}\left[\left\| \sum_{k=1}^{K} G_k N_k^t \right\|^2 \right] \right\} \\
		&\leq \mathbb{E}[F(\boldsymbol{w}^{[t]})] - \lambda \left\| \nabla F(\boldsymbol{w}^{[t]}) \right\|^2  \\
		& \quad +\frac{L\lambda^2}{2} \sum_{k=1}^{K} G_k^2 \mathbb{E}\left[ \left\| \nabla F_k(\boldsymbol{w}^{[t]}) \right\|^2 \right] + \frac{L\lambda^2}{2} \mathbb{E}\left[ \left\| \sum_{k=1}^{K} G_k N_k^t \right\|^2 \right],
	\end{aligned}
\end{equation}
where the first inequality follows from the Taylor expansion, and the second inequality follows from the Cauchy-Schwarz inequality.

From assumption \textbf{A4}, we have:
\begin{equation}
 	\frac{L}{2} \lambda^2 \cdot \sum_{k=1}^{K} G_k^2 \, \mathbb{E}\left[\left\| \nabla F_k(\boldsymbol{w}^{[t]}) \right\|^2 \right] \leq \frac{L}{2} \lambda^2 \delta^2 \sum_{k=1}^{K} G_k^2 \, \left\| \nabla F(\boldsymbol{w}^{[t]}) \right\|^2,
\end{equation}
therefore,
\begin{equation}
	\begin{aligned}
		&\mathbb{E}[F(\boldsymbol{w}^{[t+1]})] \\
		&\leq \mathbb{E}[F(\boldsymbol{w}^{[t]})] - \lambda \left\| \nabla F(\boldsymbol{w}^{[t]}) \right\|^2  \\
		& \quad +\frac{L}{2} \lambda^2 \delta^2 \sum_{k=1}^{K} G_k^2 \, \left\| \nabla F(\boldsymbol{w}^{[t]}) \right\|^2 + \frac{L}{2} \lambda^2 \cdot \mathbb{E}\left[ \left\| \sum_{k=1}^{K} G_k N_k^t \right\|^2 \right] \\
		&= \mathbb{E}[F(\boldsymbol{w}^{[t]})] + \left( \frac{L \lambda^2 \delta^2}{2} \sum_{k=1}^{K} G_k^2 - \lambda \right) \left\| \nabla F(\boldsymbol{w}^{[t]}) \right\|^2  \\
		& \quad +\frac{L}{2} \lambda^2 \cdot \mathbb{E}\left[ \left\| \sum_{k=1}^{K} G_k N_k^t \right\|^2 \right].
	\end{aligned}
\end{equation}

From assumptions \textbf{A1} and \textbf{A3}, we have $\left\| \nabla F(\boldsymbol{w}^{[t]}) \right\|^2 \geq 2\mu [F(\boldsymbol{w}^{[t]}) - F(\boldsymbol{w}^*)]$, thus:
\begin{equation}
	\begin{aligned}
		&\mathbb{E}[F(\boldsymbol{w}^{[t+1]}) - F(\boldsymbol{w}^*)]\\
		&\leq \mathbb{E}[F(\boldsymbol{w}^{[t]}) - F(\boldsymbol{w}^*)] \\
		& \quad +\left( \frac{L \lambda^2 \delta^2}{2} \sum_{k=1}^{K} G_k^2 - \lambda \right) (2\mu [F(\boldsymbol{w}^{[t]}) - F(\boldsymbol{w}^*)])  \\
		& \quad +\frac{L}{2} \lambda^2 \cdot \mathbb{E}\left[ \left\| \sum_{k=1}^{K} G_k N_k^t \right\|^2 \right] \\
		&= \mathbb{E}[F(\boldsymbol{w}^{[t]}) - F(\boldsymbol{w}^*)] \left(1 + \mu L \lambda^2 \delta^2 \sum_{k=1}^{K} G_k^2 - 2 \lambda \mu \right)  \\
		& \quad +\frac{L}{2} \lambda^2 \cdot \mathbb{E}\left[ \left\| \sum_{k=1}^{K} G_k N_k^t \right\|^2 \right],
	\end{aligned}
\end{equation}
where $A = 1 + \mu L \lambda^2 \delta^2 \sum_{k=1}^{K} G_k^2 - 2 \lambda \mu $, and $\mathbb{E}\left[ \left\| \sum_{k=1}^{K} G_k N_k^t \right\|^2 \right]$ can be divided into two parts:

(1) When $t \leq T_{{th}}$: the noise introduced by the system includes two components: artificial noise calculated based on the proposed LAPA strategy and inherent channel noise. Thus,
\begin{equation}
\begin{aligned}
    	&\mathbb{E}\left[\left\| \sum_{k=1}^{K} G_k N_k^t \right\|^2 \right] \\
        &= \sum_{k=1}^{K} G_k^2 \left( \frac{\Delta s_k^2 \cdot 2 \ln(1.25/\delta_{{dp}})}{(\epsilon_k^{[t]})^2} + \frac{\sigma_{n_0}^2}{\| \boldsymbol{h}_k \|^2 p_k^2} \right).
\end{aligned}
\end{equation}

At this point, if the FL system runs for $T_{{th}}$ rounds, we obtain:
\begin{equation}
	\begin{split}
		&\mathbb{E}[F(\boldsymbol{w}^{[t+1]}) - F(\boldsymbol{w}^*)] \\
		&\leq A^{T_{{th}}} \mathbb{E}[F(\boldsymbol{w}^{[0]}) - F(\boldsymbol{w}^*)] + \frac{1 - A^{T_{{th}}}}{1 - A} \\
		& \quad \cdot \frac{L \lambda^2}{2} \sum_{k=1}^{K} G_k^2 \left( \frac{\Delta s_k^2 \cdot 2 \ln(1.25/\delta_{{dp}})}{(\epsilon_k^{[t]})^2} + \frac{\sigma_{n_0}^2}{\| \boldsymbol{h}_k \|^2 p_k^2} \right).
	\end{split}
\end{equation}

(2) When $t > T_{{th}}$: the system stops adding artificial noise, so only channel noise remains, and:
\begin{equation}
	\mathbb{E}\left[\left\| \sum_{k=1}^{K} G_k N_k^t \right\|^2 \right] = \sum_{k=1}^{K} G_k^2 \cdot \frac{\sigma_{n_0}^2}{\| \boldsymbol{h}_k \|^2 p_k^2}.
\end{equation}

At this point, letting the FL system run for the remaining rounds gives:
\begin{equation}
	\begin{split}
		&\mathbb{E}[F(\boldsymbol{w}^{[t+1]}) - F(\boldsymbol{w}^*)] \\
		&\leq A^{T - T_{{th}}} \mathbb{E}[F(\boldsymbol{w}^{[0]}) - F(\boldsymbol{w}^*)]  \\
		& \quad +\frac{1 - A^{T - T_{{th}}}}{1 - A} \cdot \frac{L \lambda^2}{2} \sum_{k=1}^{K} G_k^2 \cdot \frac{\sigma_{n_0}^2}{\| \boldsymbol{h}_k \|^2 p_k^2}.
	\end{split}
\end{equation}

In summary, when the FL process runs for $T$ rounds, we have:
\begin{equation}
	\begin{aligned}
		&\mathbb{E}[F(\boldsymbol{w}^{[t+1]}) - F(\boldsymbol{w}^*)] \\
		&\leq A^T \mathbb{E}[F(\boldsymbol{w}^{[0]}) - F(\boldsymbol{w}^*)] 
		+ \frac{L \lambda^2}{2} \cdot \frac{1 - A^T}{1 - A} \sum_{k=1}^{K} G_k^2 \cdot \frac{\sigma_{n_0}^2}{\| \boldsymbol{h}_k \|^2 p_k^2} \\
		&\quad + \frac{L \lambda^2}{2} \sum_{m=0}^{T_{{th}}} A^m \left[ \sum_{k=1}^{K} \frac{G_k^2 \Delta s_k^2 \cdot 2 \ln(1.25/\delta_{{dp}})}{(\epsilon_k^{[m]})^2} \right].
	\end{aligned}
\end{equation}

Proof completed.

\bibliographystyle{IEEEtranTIE}
\bibliography{references}\ 

\vfill

\end{document}